\documentclass{article}

\PassOptionsToPackage{numbers,sort&compress}{natbib}
 \usepackage[main, final]{neurips_2026}
 \usepackage{subcaption}

\setcitestyle{numbers,square,comma}

\usepackage[utf8]{inputenc} 
\usepackage[T1]{fontenc}    
\usepackage{hyperref}       
\usepackage{url}            
\usepackage{booktabs}       
\usepackage{amsfonts}       
\usepackage{nicefrac}       
\usepackage{microtype}      
\usepackage{xcolor}         
\usepackage{algorithm}
\usepackage{algpseudocode}
\usepackage{amssymb}
\usepackage{multirow}

\usepackage[table]{xcolor}
\usepackage{booktabs}

\usepackage{booktabs}
\usepackage{multirow}
\usepackage{graphicx}
\usepackage[table]{xcolor}
\usepackage{titletoc}

\definecolor{sectionblue}{RGB}{223,235,255}
\definecolor{dfree}{RGB}{245,248,255}
\definecolor{standard}{RGB}{248,248,248}
\definecolor{copehl}{RGB}{240,250,243}
\definecolor{copegreen}{RGB}{0,110,55}

\usepackage[dvipsnames]{xcolor}
\usepackage[table]{xcolor}
\usepackage{multirow}
\definecolor{softgold}{RGB}{218,165,32}
\definecolor{softgoldrow}{RGB}{255,248,220}

\newcommand{\slowblk}[1]{\textcolor{BrickRed}{#1}}
\newcommand{\slerpblk}[1]{\textcolor{RoyalBlue}{#1}}
\newcommand{\memblk}[1]{\textcolor{ForestGreen}{#1}}
\newcommand{\corrblk}[1]{\textcolor{Magenta}{#1}}
\newcommand{\proxblk}[1]{\textcolor{TealBlue}{#1}}
\newcommand{\forgetblk}[1]{\textcolor{Orange}{#1}}

\usepackage{xspace}
\newcommand{\method}{\textsc{FOGO}\xspace}

\title{FOGO: Forgetting-aware \\ Orthogonalization Optimizer}

\author{%
  Toan Nguyen \\
  UNSW Sydney, Australia\\
  DataCruiser, Australia \\
  \texttt{toan.nguyen@unsw.edu.au} \\
  \And
  Yang Liu \\
  UNSW Sydney, Australia\\
  \texttt{yang.liu39@student.unsw.edu.au} \\
  \And
  Trung Le \\
  Monash University, Australia\\
  \texttt{trunglm@monash.edu} \\
  \And
  Celso De Melo \\
  DEVCOM Army Research Laboratory, USA\\
  \texttt{celso.miguel.de.melo@gmail.com} \\
  \And
  Flora D. Salim\thanks{Corresponding Author.} \\
  UNSW Sydney, Australia\\
  DataCruiser, Australia \\
  \texttt{flora.salim@unsw.edu.au} \\
}

\begin{document}

\maketitle

\begin{abstract}

We argue that forgetting is not confined to continual learning but is a general optimization phenomenon: during standard training, dominant mini-batch gradients suppress rare but useful update directions, causing short-term forgetting at every step. When such knowledge is never revisited, these losses compound into long-term forgetting, the classical failure mode of continual learning. We introduce \method{}, a scalable optimizer that continuously detects and resolves gradient interference across both regimes. \method{} spectrally orthogonalizes momentum updates to prevent dominant directions from monopolizing optimization, then stores representative past directions in a compact codebook memory built on random projection, where pairwise distances are approximately preserved in a low-dimensional space. At each step, it resolves conflicts between the current update and the stored directions with a lightweight orthogonal correction and lifts the result back through a proximal step, with minimal overhead and without storing raw data. For replay-free task sequences, the same update stream can also be summarized by its second-order statistics, giving a full-capacity variant that outperforms a strong gradient-projection baseline at comparable memory, without storing any sample. Across class-imbalanced classification, continual visual learning under domain and class shifts, continual fine-tuning of LLaVA-7B, and GPT-2 pretraining, \method{} consistently improves convergence and knowledge retention over Adam and Muon. Code is available at \url{https://github.com/cruiseresearchgroup/FOGO}.


\end{abstract}

\section{Introduction}
\label{sec:Intro}

\providecommand{\rev}[1]{\textcolor{red}{#1}}%

Deep learning has achieved remarkable progress across vision~\citep{he2016deep}, language~{\citep{radford2019language,touvron2023llama}}, and multimodal tasks~{\citep{liu2023visual}}, yet its dominant training paradigm struggles when knowledge must be retained and accumulated over time. Humans naturally integrate new experiences without overwriting old ones; neural networks, optimized under stationary regimes, lack this ability and degrade under distribution shift. Continual learning formalizes this challenge~\citep{de2021continual}, {but it has largely been studied separately from standard training, with its own benchmarks, methods, and metrics}.

Recent work has begun to narrow this divide: \citet{abel2023a} reinterpret continual learning as continual adaptation, while gradient-interference analyses show that forgetting arises when new updates overwrite parameter directions useful for prior knowledge~\citep{farajtabar2020orthogonal,riemer2019learning}. We push this further and argue that \emph{all} learning is a continual process of resolving interference between incoming and existing knowledge. Under this lens, standard training and task-sequential continual learning are not separate problems but different regimes of the same phenomenon. In standard training, dominant mini-batch gradients can suppress rare but useful directions, causing \emph{short-term forgetting} at each optimization step. When such knowledge is not revisited, these losses accumulate into \emph{long-term forgetting}{, the} classical failure mode of continual learning. We validate this through controlled experiments on class-imbalanced CIFAR-10: short-term forgetting {occurs and degrades accuracy on} underrepresented classes. Random batching mitigates it in common balanced settings{, which explains why it has been overlooked,} but fails under imbalance or distribution shift.

{Motivated by this view, we} introduce \method{}, a scalable optimizer that continuously detects and resolves gradient conflicts across both forgetting regimes. \method{} combines two mechanisms. First, it spectrally orthogonalizes momentum updates, preventing dominant directions from monopolizing optimization and exposing rare but informative components. Second, it stores useful update directions in a compact codebook memory. {Following recent work on measuring data diversity~\citep{jung2025prismatic}, \method{} uses random projection to map} high-dimensional updates to a low-dimensional space where pairwise distances are approximately preserved~\citep{johnson1984extensions}, maintains representative directions as centroids, and resolves conflicts via lightweight orthogonal correction before lifting the result back through a proximal step. This yields an optimizer-level memory that operates entirely in projected space, keeping per-step overhead minimal. {For replay-free task sequences, the same update stream can also be summarized at second order, giving a full-capacity long-term memory that outperforms GPM~\citep{saha2021gradient} under its own protocol.}

We evaluate \method{} {on} class-imbalanced classification, continual visual learning under domain and class shifts, {continual fine-tuning of LLaVA-7B} on visual question answering, and GPT-2 pretraining. Across all settings, \method{} consistently improves retention and convergence, outperforming Adam~{\citep{kingma2014adam}} and recent matrix-aware alternatives such as Muon~{\citep{jordan2024muon}}. These results suggest that forgetting is not only a challenge of task-sequential learning but a general optimization problem{, and} that directional memory offers a practical route toward resolving it.

\section{Related Work}
\label{sec:Related}

\providecommand{\rev}[1]{\textcolor{red}{#1}}%
\subsection{Matrix-aware and Orthogonalized Optimizers}
Matrix-aware optimizers have re-emerged as strong alternatives to diagonal adaptive methods~\citep{si2025adamuon,liu2025muon}. Shampoo~\citep{gupta2018shampoo} captures richer geometry via structured tensor preconditioning, while Muon~\citep{jordan2024muon} simplifies this idea by orthogonalizing momentum updates with Newton--Schulz iterations, yielding strong LLM pretraining results~\citep{liu2025muon}. Recent work studies Muon through spectral-norm geometry~\citep{bernstein2024old} and {extends it with adaptive variants}~\citep{si2025adamuon}. \method{} goes beyond using orthogonalized updates for conditioning: it stores useful update directions in efficient codebook memories to mitigate both short-term and long-term forgetting in standard and continual learning.
\subsection{Continual Learning}
Continual learning is commonly formulated as a task-sequential problem, where forgetting is measured by performance degradation on previous tasks. A complementary view attributes forgetting to gradient interference~\citep{farajtabar2020orthogonal,riemer2019learning,lopez2017gradient}, while \citet{abel2023a} reframe learning as endless adaptation rather than convergence to a fixed solution. We draw on both: \method{} treats standard training and task-sequential continual learning as regimes of a single interference problem, distinguished by the source, frequency, and timescale of conflicting gradients.
Among existing methods~\citep{rusu2016progressive,buzzega2020dark}, gradient-constrained ones are closest to ours. GEM and A-GEM~\citep{lopez2017gradient,chaudhry2018efficient} constrain updates via reference gradients from stored examples, tying anti-forgetting to raw data retention. OGD~\citep{farajtabar2020orthogonal} and GPM~\citep{saha2021gradient} avoid stored data by projecting onto orthogonal subspaces, but rely on {full-dimensional} gradient bases or SVD over high-dimensional activations. O-LoRA~\citep{wang2023orthogonal} and InfLoRA~\citep{liang2024inflora} instead bound interference architecturally, freezing the adapter branches of previous tasks and adding a near-orthogonal branch per task, which needs task boundaries and a parameter-efficient architecture. Recent optimizers such as UPGD~\citep{elsayed2024addressing} and spectral-orthogonalization methods~\citep{behrouz2025nested,han2026fire} address non-stationarity more directly, yet none persistently tracks past gradient geometry to detect inter-task conflicts. \method{} closes this gap with a compact gradient memory via random projection, which {approximately} preserves pairwise distances~\citep{johnson1984extensions}, enabling per-step interference detection without raw data or full-rank decompositions. {As the projected dimension does not depend on the parameter dimension}, \method{} scales naturally to the billion-parameter regime demanded by continual adaptation of large pre-trained models~\citep{wang2023orthogonal,huai2025cl}.

\section{Background}
\label{sec:Background}

\subsection{Muon and Orthogonalization-Based Optimizers}
 
Given a momentum matrix $B_t \in \mathbb{R}^{m\times n}$ with compact SVD $B_t = U_t \Sigma_t V_t^\top$, the \emph{orthogonalized update} is its polar factor $\mathcal{O}(B_t) := U_t V_t^\top$, which removes singular-value scaling while preserving directional structure. In practice, $\mathcal{O}(B_t)$ is efficiently approximated by Newton--Schulz iterations~\citep{higham2008functions,bernstein2024old}. Muon~\citep{jordan2024muon} applies this idea to matrix-valued parameters: given weight $W_t \in \mathbb{R}^{m\times n}$ and gradient $G_t = \nabla_W \mathcal{L}(W_t)$,
\begin{equation}
B_t = \mu B_{t-1} + G_t, \qquad O_t = \mathrm{Newton\mathrm{-}Schulz}(B_t, T), \qquad W_{t+1} = W_t - \eta\, O_t,
\end{equation}
where $B_0 = \mathbf{0}$ and $\mu, \eta \in (0,1)$. By flattening the singular spectrum, orthogonalization prevents a few dominant modes from dictating the update, letting weaker but coherent directions contribute more strongly. This perspective extends to other orthogonalization-based optimizers~\citep{gupta2018shampoo,behrouz2025nested} and motivates our continual optimizer for mitigating forgetting.
 
\subsection{Orthogonalization in Continual Learning}
 
Orthogonal gradient methods project each update onto the orthogonal complement of past-task directions to prevent interference~\citep{farajtabar2020orthogonal,wang2023orthogonal}. OGD~\citep{farajtabar2020orthogonal} maintains a set $S$ of past gradient directions and projects the current gradient as:
\begin{equation}
\tilde{G} = G - \sum_{G' \in S} \frac{\langle G, G' \rangle}{\langle G', G' \rangle}\, G'.
\end{equation}
A key limitation is scalability: $|S|$ grows with the number of tasks, making both storage and per-step projection prohibitive for large models.
 
\subsection{Random Projection of Gradients}
\label{subsec:random-projection}
 
Operating directly in parameter space is expensive. Following recent work on gradient-space diversity measurement~\citep{jung2025prismatic,friedman2023the}, we apply a Rademacher random projection $\Pi \in \mathbb{R}^{n \times d}$, $\Pi_{ij} \sim \mathcal{U}(\{-1,1\})$, to each row of a layer's update, with $d \ll n$. As a Johnson--Lindenstrauss transform~\citep{johnson1984extensions}, this preserves pairwise distances up to $(1 \pm \epsilon)$ distortion with high probability when $d = \mathcal{O}(\epsilon^{-2} \log N)$, where $N$ counts the rows and centroids being compared. The projected rows yield a compact, geometry-preserving embedding of the update distribution, whose structure reveals representative update directions. We provide extended background in Appendix~\ref{sec:extended_background}.

\section{Method}
\label{sec:Method}

\subsection{Why Learning is Prone to Forgetting}
 
\paragraph{Motivation.}
Forgetting is typically framed as a consequence of task boundaries or distribution shifts~\citep{wang2024comprehensive}. We argue that it can arise even in standard training. Under-represented data subsets can produce update directions that conflict with dominant updates, which can gradually overwrite what the model learned from those subsets. \emph{The same mechanism can lead to forgetting on two timescales}. In standard training, forgetting can be short-lived because later batches may revisit these subsets. In continual learning, it can persist when earlier data remain unavailable.

\paragraph{Dominant subspace.}
The cosine between the incoming gradient $G_t$ and the momentum $B_{t-1}$ is a natural measure of interference, but it is too coarse: the two can agree overall while $G_t$ still erases a few directions that matter for under-represented data. We therefore look inside the momentum. Transformer updates often have concentrated singular spectra~\citep{shen2025convergence,hu2022lora}, so most of the energy of $B_t$ lives in a small dominant subspace; let $U_t^{(r)}$ and $V_t^{(r)}$ hold its top-$r$ singular vectors.
 
\begin{figure}[t]
  \centering
  \includegraphics[width=\linewidth]{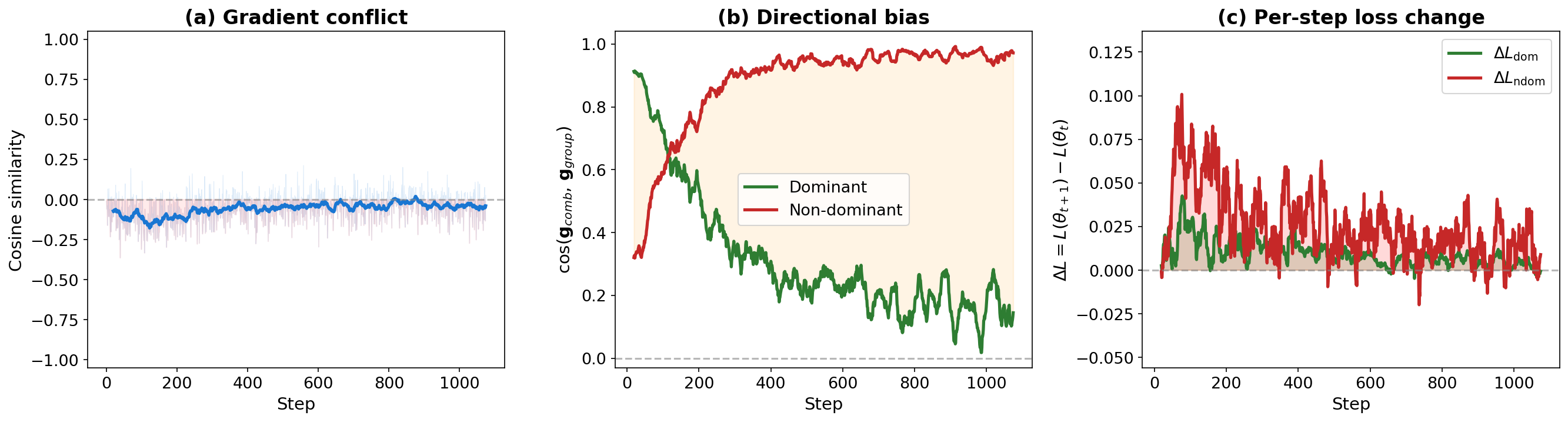}
  \caption{%
    \textbf{Short-term forgetting under class imbalance}
    (CIFAR-10, ResNet-18, 5 rare classes at 10\% data).
    \textbf{(a)}~Dominant and rare gradients maintain negative cosine similarity throughout training.
    \textbf{(b)}~The combined update initially aligns with the dominant group; after dominant loss saturates (${\sim}$step 200), the bias reverses.
    \textbf{(c)}~Per-step change in validation loss on a held-out probe. While the update serves the dominant group, each step raises the rare group's loss, by up to $+0.10$, as Corollary~2 predicts.
  }
  \label{fig:short-term-forgetting}
\end{figure}
 
\paragraph{Theorem 1 (Suppression of off-subspace directions).}
Let $\mathcal P_t^{(r)}(X) = U_t^{(r)}U_t^{(r)\top} X V_t^{(r)}V_t^{(r)\top}$ be the two-sided projection onto the top-$r$ subspace of $B_t$, with $\mathcal P_t^{(r)\perp}(X) = X - \mathcal P_t^{(r)}(X)$. Define $R_t := \|\mathcal P_t^{(r)\perp}(B_t)\|_F$ and $\varepsilon_t := \|\mathcal P_t^{(r)\perp}(G_t)\|_F$. Under momentum update $B_t = \mu B_{t-1} + G_t$ with $\mu \in [0,1)$, if the dominant subspace drifts slowly, $\|\mathcal P_t^{(r)} - \mathcal P_{t-1}^{(r)}\|_{F \to F} \le \delta_t$, then
\[
R_t \le \mu R_{t-1} + \mu\delta_t\|B_{t-1}\|_F + \varepsilon_t.
\]
Off-subspace mass grows only through subspace drift ($\mu\delta_t\|B_{t-1}\|_F$) and newly injected off-subspace energy ($\varepsilon_t$). When both are small, the residual contracts while dominant directions accumulate. Theorem~1 establishes suppression; the two corollaries below turn suppression into forgetting (proofs in Appendix~\ref{sect:theoretical_results_appendix}).

\paragraph{Corollary 2 (One-step degradation).}
Let $\mathcal L_S$ be the $\beta$-smooth loss of a data subset $S$, such as a rare group or a previous task. The step $\theta_{t+1}=\theta_t-\eta u_t$ strictly increases $\mathcal L_S$ whenever $\langle\nabla\mathcal L_S(\theta_t),u_t\rangle<-\frac{\beta\eta}{2}\|u_t\|_2^2$. By Theorem~1 the update carries little mass along the directions that serve $S$; once its dominant components point against $\nabla\mathcal L_S$, the condition is met and the loss on $S$ rises within a single step. This is not a mini-batch artifact: take a dominant group and a rare group with proportions $p_D,p_R$ and gradients $g_D,g_R$; even exact full-batch descent raises the rare group's loss, for small enough $\eta$, as soon as $\langle g_R,g_D\rangle<-\frac{p_R}{p_D}\|g_R\|_2^2$. Since $p_R/p_D$ is small, a mild conflict between the two gradients already does this.

\paragraph{Corollary 3 (Persistence).}
Fix a unit direction $H\in\mathbb R^{m\times n}$ and let $b_t:=\langle B_t,H\rangle_F$ be its coefficient in the momentum. If $\kappa_-\le\mathbb E[\langle G_t,H\rangle_F\mid\mathcal F_{t-1}]\le\kappa_+$ for all $t>t_0$, then for all $t\ge t_0$, $$\mu^{t-t_0}b_{t_0}+\kappa_-\frac{1-\mu^{t-t_0}}{1-\mu}\;\le\;\mathbb E[b_t\mid\mathcal F_{t_0}]\;\le\;\mu^{t-t_0}b_{t_0}+\kappa_+\frac{1-\mu^{t-t_0}}{1-\mu}.$$ Take $H$ useful for $S$. When later batches revisit $S$, $\kappa_->0$ and the coefficient is held above $\kappa_-/(1-\mu)$, so the damage of Corollary~2 is repaired; under imbalance this level scales with the rare-group frequency $p_R$; under task shift nothing re-injects mass along $H$, $\kappa_+\le0$, and the bound decays geometrically. This is long-term forgetting.

Fig.~\ref{fig:short-term-forgetting} verifies Corollary~2 on class-imbalanced CIFAR-10~\citep{krizhevsky2009learning}: the two group gradients stay in conflict~(a), the update follows the dominant group~(b), and the rare group's loss rises after each step~(c); details are in Appendix~\ref{appendix:short-term-forgetting}. That interference causes forgetting is the premise of GEM, A-GEM, OGD, and GPM~\citep{lopez2017gradient,chaudhry2018efficient,farajtabar2020orthogonal,saha2021gradient}; our point is how it arises: through the dominant modes of the momentum state, in standard training as much as in task sequences.

\paragraph{Two axes of interference.}
Orthogonalization only rescales the singular values of $B_t$ (Proposition~6, Appendix~\ref{subsec:appendix_orthogonalization_spectral}). It equalizes the directions still present in $B_t$, but it cannot bring back one that has faded out ($b_t=0$), and it has no way to tell a suppressed useful direction from mini-batch noise, since both appear as small singular values. Protecting past directions therefore needs state outside $B_t$ that records which directions were useful: the directional memory of Section~\ref{subsec:fogo_method}.

\subsection{\method{}: Forgetting-aware Orthogonalization Optimizer}
\label{subsec:fogo_method}

Fig.~\ref{fig:update_flow} in Appendix~\ref{sec:update_flow_appendix} shows the update flow of Muon and \method{} side by side: both start from orthogonalized momentum; \method{} adds the projected codebook, the two corrections, and the proximal lift before the weights are updated.

\subsubsection{From Orthogonalization to Directional Memory}

Orthogonalizing momentum preserves singular directions while flattening their spectrum, preventing high-energy modes from monopolizing the update and keeping weaker but useful directions visible (Proposition~6, Appendix~\ref{subsec:appendix_orthogonalization_spectral}). However, orthogonalization alone does not protect these directions over time (Corollary~3): future gradients can overwrite them, and storing raw gradients is infeasible at scale.
\method{} addresses this with a compact directional memory in a low-dimensional projected space. Using random projection~\citep{johnson1984extensions,jung2025prismatic}, it tracks representative past directions whose pairwise geometry is preserved with high probability, then corrects each orthogonalized update against this memory before applying it, mitigating conflicts at controlled overhead.



\subsubsection{Memory of Gradients}
\label{subsubsec:memory_of_gradients}

\paragraph{\slowblk{Slow--fast orthogonalized streams.}}
To capture both stable and emerging directions, \method{} maintains two momentum states at different timescales. At step $t$, given gradient $G_t = \nabla_\theta \mathcal{L}_t(\theta_{t-1}) \in \mathbb{R}^{m \times n}$:
\[
B_t^{(s)} = \mu_s B_{t-1}^{(s)} + G_t, \qquad B_t^{(f)} = \mu_f B_{t-1}^{(f)} + G_t, \qquad \mu_s > \mu_f, \quad B_t^{(s)}, B_t^{(f)} \in \mathbb{R}^{m \times n},
\]
where the slow stream captures persistent trends and the fast stream tracks recent changes. Both are orthogonalized via Newton--Schulz iterations and projected into a compact $d$-dimensional space ($d \ll n$):
\[
O_t^{(\cdot)} = \textsc{NewtonSchulz}(B_t^{(\cdot)}; K_{(\cdot)}), \qquad P_t^{(\cdot)} = O_t^{(\cdot)} \Pi \in \mathbb{R}^{m \times d},
\]
where $\Pi \in \mathbb{R}^{n \times d}$ is a fixed Rademacher random matrix with entries drawn i.i.d.\ from $\mathcal{U}(\{-1,1\})$ and scaled by $1/\sqrt d$. For each row $i$, we \emph{fuse} the projected slow and fast directions via spherical interpolation ($z'_{t,i}, z_{t,i} \in \mathbb{R}^d$):
\[
z'_{t,i} = \textsc{Slerp}(P_{t,i}^{(s)}, P_{t,i}^{(f)}; \xi), \qquad z_{t,i} = \|P_{t,i}^{(s)}\|_2 \, z'_{t,i},
\]
where $\xi$ controls interpolation strength and the slow stream sets the scale. Stacking rows gives $Z_t = [z_{t,1}; \ldots; z_{t,m}] \in \mathbb{R}^{m \times d}$. Details are in Appendix~\ref{subsec:detail_fusion}.

\paragraph{\memblk{Online directional memory.}}
\method{} maintains a spherical codebook $\mathcal{C}_t = \{c_{t,1}, \ldots, c_{t,C}\} \subset \mathbb{R}^d$ that summarizes recurring update directions. Each fused row $z_{t,i}$ is normalized to $q_{t,i} = z_{t,i} / (\|z_{t,i}\|_2 + \varepsilon)$ and assigned to its nearest centroid under cosine similarity:
\[
a_{t,i}^{\mathrm{mem}} = \arg\max_{j \in \{1,\ldots,C\}} \langle q_{t,i}, \bar{c}_{t-1,j} \rangle, \qquad \bar{c}_{t-1,j} = \frac{c_{t-1,j}}{\|c_{t-1,j}\|_2 + \varepsilon}.
\]
Assignments are accumulated into EMA sufficient statistics and flushed after each memory-guided update to obtain $\mathcal{C}_t$, with centroids re-normalized onto the unit sphere. Codebook collapse is prevented via decorrelation against nearby centroids and periodic re-seeding of inactive ones (Appendix~\ref{subsec:detail_online_directional_memory}). The codebook also gives a concrete meaning to a useful direction, without any labels: a direction that keeps showing up across steps builds a stable centroid, even if it is weak in every single batch, whereas a direction that appears once averages out of the EMA statistics and its centroid is eventually re-seeded. The corrections below only act on updates that conflict with such recurring directions.

\subsubsection{Addressing Short-Term and Long-Term Forgetting}
\label{subsubsec:short_long_term}
 
\method{} compares each orthogonalized update against its codebook of gradient centroids and corrects directions that interfere with stored memory, performing memory-guided updates that preserve plasticity while suppressing destructive interference.

\paragraph{\corrblk{Short-term angular-band filtering.}}
Short-term forgetting arises not only from anti-aligned updates but also from repeatedly reinforcing dominant local directions. \method{} treats a near-orthogonal angular band around each pre-update centroid as locally safe and corrects updates that fall outside it.
 
For $q_{t,i} = z_{t,i} / (\|z_{t,i}\|_2 + \varepsilon)$, we partition centroids into over-aligned and conflicting sets:
\[
\mathcal{H}_{t,i}^{\mathrm{st}} = \{j : \langle q_{t,i}, \bar{c}_{t-1,j} \rangle > \epsilon_{\mathrm{st}}\}, \qquad \mathcal{C}_{t,i}^{\mathrm{st}} = \{j : \langle q_{t,i}, \bar{c}_{t-1,j} \rangle < -\epsilon_{\mathrm{st}}\}.
\]
We form the projected components and subtract them:
\[
z_{t,i}^{\mathrm{hi}} = \sum_{j \in \mathcal{H}_{t,i}^{\mathrm{st}}} \alpha_{t,ij}^{+} \langle z_{t,i}, \bar{c}_{t-1,j} \rangle \bar{c}_{t-1,j}, \qquad z_{t,i}^{\mathrm{conf}} = \sum_{j \in \mathcal{C}_{t,i}^{\mathrm{st}}} \alpha_{t,ij}^{-} \langle z_{t,i}, \bar{c}_{t-1,j} \rangle \bar{c}_{t-1,j},
\]
\[
\tilde{z}_{t,i} = z_{t,i} - \gamma_{\mathrm{hi}}\, z_{t,i}^{\mathrm{hi}} - \gamma_{\mathrm{st},t}\, z_{t,i}^{\mathrm{conf}}.
\]
Conflicting components are explicitly removed, while over-aligned components are only mildly deflated ($\gamma_{\mathrm{hi}}$ small) as an exploration regularizer against collapse onto dominant directions. The coefficient $\gamma_{\mathrm{st},t}$ adapts to batch-level energy outside the safe band. Weighting and scheduling details are in Appendix~\ref{subsec:detail_short_term_filtering}.

\paragraph{\forgetblk{Long-term near-orthogonal protection.}}
After completing task $\mathcal{T}_\tau$, \method{} freezes important centroids $\{\bar{c}_{\mathcal{T}_\tau, k}\}_{k=1}^{K_\tau}$ with importance weights $\{\nu_{\mathcal{T}_\tau, k}\}_{k=1}^{K_\tau}$ as long-term memory of sensitive past-task directions. At step $t$ of the current task, updates should remain nearly orthogonal to these directions. Given the short-term corrected $\tilde{z}_{t,i}$ with $\tilde{q}_{t,i} = \tilde{z}_{t,i} / (\|\tilde{z}_{t,i}\|_2 + \varepsilon)$, we activate protection for centroids not sufficiently orthogonal:
\[
\mathcal{B}_{t,i}^{\mathrm{lt}} = \bigl\{(\mathcal{T}_\tau, k) \in \mathcal{A}_t^{\mathrm{lt}} : |\langle \tilde{q}_{t,i}, \bar{c}_{\mathcal{T}_\tau, k} \rangle| > \epsilon_{\mathrm{lt}}\bigr\},
\]
where $\mathcal{A}_t^{\mathrm{lt}}$ denotes available frozen centroids and $\epsilon_{\mathrm{lt}} < \epsilon_{\mathrm{st}}$ enforces stricter protection. The long-term correction subtracts the importance-weighted protected component:
\[
z_{t,i}^{\mathrm{lt}} = \sum_{(\mathcal{T}_\tau, k) \in \mathcal{B}_{t,i}^{\mathrm{lt}}} \nu_{\mathcal{T}_\tau, k} \langle \tilde{z}_{t,i}, \bar{c}_{\mathcal{T}_\tau, k} \rangle \bar{c}_{\mathcal{T}_\tau, k}, \qquad \bar{z}_{t,i} = \tilde{z}_{t,i} - \gamma_{\mathrm{lt}}\, z_{t,i}^{\mathrm{lt}}.
\]
Long-term memory thus acts as an importance-weighted orthogonality constraint, suppressing sensitive directions while preserving free subspaces for adaptation. Details on centroid selection, candidate sampling, and importance weighting are in Appendix~\ref{subsec:detail_long_term_protection}.

\paragraph{\forgetblk{Long-term memory at two capacities.}}
\label{subsubsec:two_capacities}
Frozen centroids protect $K_\tau$ individual directions per task (Theorem~4). This suffices when later data keep touching the old directions (replay, pretrained backbones, imbalance): a direction that drifts is pulled back. In a replay-free task sequence nothing pulls it back (Corollary~3, $\kappa_+\le0$), and the whole subspace of a past task must be protected. For this regime \method{} keeps a second-order memory of the same update stream, without projection ($d=n$). Within a task, each layer accumulates the covariance of its filtered rows $\tilde z_{t,i}\in\mathbb R^n$,
\[
\Psi_t=\rho\,\Psi_{t-1}+\sum_{i=1}^{m}\tilde z_{t,i}\tilde z_{t,i}^{\top}\in\mathbb R^{n\times n}.
\]
At the task boundary, the top eigenvectors of $\Psi$ are added to an orthonormal basis $Q\in\mathbb R^{n\times r}$ with $r\le n/2$. In later tasks $Q$ replaces the centroids, $\Omega=QQ^\top$ with $\gamma_{\mathrm{lt}}=1$: protection subtracts from each row its projection onto $\mathrm{span}(Q)$,
\[
\bar z_{t,i}=(I-QQ^\top)\,\tilde z_{t,i},
\]
and the same subtraction is applied once more to the lifted update $\hat O_t$, so what reaches the weights is exactly orthogonal to $\mathrm{span}(Q)$. Momentum, short-term filtering, and lifting are unchanged. 

\subsection{Projection Back to Original Space and Parameter Update}

Memory corrections are computed in the projected space. Since $\Pi \in \mathbb{R}^{n \times d}$ is not invertible, we lift the protected direction back via a proximal least-squares problem. For row $i$, let $\Delta z_{t,i} = \bar{z}_{t,i} - P_{t,i}^{(s)} \in \mathbb{R}^d$ be the projected correction relative to the slow stream. We seek the minimal high-dimensional correction consistent with $\Delta z_{t,i}$:
\[
\delta_{t,i}^{\mathrm{prox}} = \arg\min_{\delta \in \mathbb{R}^n} \left\{ \frac{1}{2}\|\delta\|_2^2 + \lambda\|\Pi^\top \delta - \Delta z_{t,i}\|_2^2 \right\}.
\]
With $M = \Pi^\top \Pi$, the closed-form solution and lifted protected update are
\[
u_{t,i}^{\mathrm{prox}} = (I + 2\lambda M)^{-1}(2\lambda \Delta z_{t,i}), \qquad \delta_{t,i}^{\mathrm{prox}} = \Pi\, u_{t,i}^{\mathrm{prox}}, \qquad \hat{O}_{t,i} = O_{t,i}^{(s)} + \delta_{t,i}^{\mathrm{prox}}.
\]
Stacking all rows gives $\hat{O}_t \in \mathbb{R}^{m \times n}$. The parameter update is
\[
\theta_t = (1 - \eta\,\mathrm{wd})\,\theta_{t-1} - \eta\,\gamma_t\,\hat{O}_t, \qquad \gamma_t = \sigma \frac{\sqrt{mn}}{\|\hat{O}_t\|_F + \varepsilon},
\]
where $\mathrm{wd}$ is the weight decay coefficient, $\varepsilon = 10^{-8}$, and $\gamma_t$ follows the scaling of~\citet{si2025adamuon}. Overall, \method{} combines dual-stream orthogonalized updates with multiscale directional memory: geodesic fusion preserves plasticity, the online codebook filters short-term interference, and frozen centroids impose stricter long-term protection. The proximal step maps these corrections back to parameter space with controlled distortion.

\subsection{Theoretical Analysis}

\paragraph{Theorem 4 (Memory protection preserves learned directions).}
\label{thm:learned_knowledge_interference}
Let $x\in\mathbb R^{d}$ be an update row before proximal lifting, and let $\Omega=\sum_{a\in\mathcal I}\nu_a c_a c_a^\top$ be the memory operator built from the active unit directions $c_a$ with weights $\nu_a\ge0$. The protected update is $x^{+}=(I-\gamma\,\Omega)\,x$. If $\Omega\neq0$ and $0\le\gamma\le1/\lambda_{\max}(\Omega)$, then
\[
\|\Omega^{1/2}x^{+}\|_2\le\|\Omega^{1/2}x\|_2
\qquad\mathrm{and}\qquad
x^{+}=x\ \ \mathrm{whenever~} \Omega x=0 .
\]
That is, protection shrinks the update along the stored directions and leaves every other direction untouched. The proof (Appendix~\ref{sect:theoretical_results_appendix}) uses only that $\Omega$ is positive semidefinite, so the statement holds in any dimension and for both memories of Section~\ref{subsubsec:two_capacities}: for frozen centroids, the free subspace is $\mathrm{Null}(\Omega)$; for the second-order basis, $\Omega=QQ^\top$ has $\lambda_{\max}=1$, and $\gamma_{\mathrm{lt}}=1$ removes the protected component exactly. For centroids the implementation rescales the weights so that $\gamma_{\mathrm{lt}}\lambda_{\max}(\Omega)\le1$ (Appendix~\ref{subsec:detail_long_term_protection}).

\section{Experiments}
\label{sec:Experiments}

\subsection{Class-Imbalanced Learning}
\label{subsection:class-imbalanced-learning}

\paragraph{Setup.}
We introduce a controlled class-imbalanced CIFAR-10 benchmark to probe \emph{short-term forgetting}. ResNet-18~\cite{he2016deep} is trained from scratch for 50 epochs, with five \emph{rare} classes retaining only 10\% of their training samples. This creates sparse rare-class gradients that are persistently dominated by majority-class updates. We report rare- and non-rare-class accuracy separately on the standard test set, measuring rare-class forgetting as the best-so-far accuracy drop. Details are in Appendix~\ref{sec:more_exps_settings_appendix}.

\begin{figure}[t]
    \centering
    \includegraphics[width=0.80\linewidth]{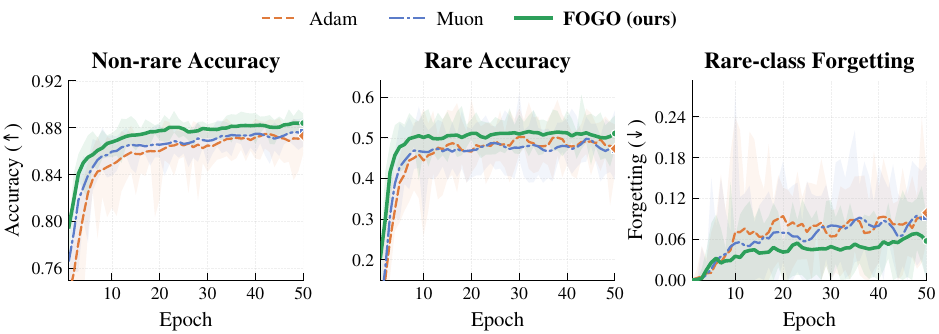}
    \vspace{0.3em}
    \includegraphics[width=0.80\linewidth]{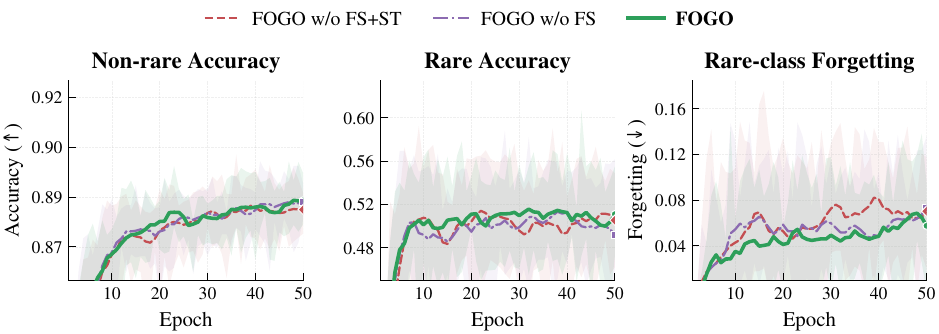}

    \caption{\textbf{Class-imbalanced CIFAR-10.}
   \textbf{Top:} \method{} outperforms Adam and Muon under class-imbalanced training, improving rare-class accuracy and reducing rare-class forgetting while preserving non-rare accuracy.
    \textbf{Bottom:} Removing slow--fast fusion (w/o FS) and short-term filtering (w/o FS+ST) progressively increases rare-class forgetting.
    Curves show the mean over ten seeds, with shaded regions denoting one standard deviation.}
    \label{fig:class_imbalanced_cifar10}
\end{figure}

\paragraph{Results and Ablation.}
As shown in Fig.~\ref{fig:class_imbalanced_cifar10}, \method{} improves both rare- and non-rare-class accuracy over Adam and Muon by roughly 2--3 percentage points while reducing rare-class forgetting by about 6 points, indicating that its correction preserves low-frequency learning signals without trading off frequent-class performance. The ablation shows that removing fast--slow fusion causes mild degradation, whereas removing both fusion and short-term filtering substantially increases forgetting—confirming short-term filtering as the primary contributor, with fusion providing additional gain. Appendix~\ref{subsec:training_curve_appendix} shows that \method{} also achieves smoother and faster convergence.

\subsection{Continual Visual Learning}

\paragraph{Setup.} We evaluate \method{} on continual visual learning in two regimes. In \emph{domain-incremental learning (Domain-IL)}, we use PACS~\citep{li2017deeper}, where seven classes appear sequentially across four domains (Photo, Art Painting, Cartoon, Sketch). We train an ImageNet-pretrained ResNet-18 with fixed batch-normalization statistics and test \method{} both without replay and with ER~\citep{chaudhry2019tiny}, comparing against plasticity-oriented optimizers including CBP~\citep{dohare2024loss} and FIRE~\citep{han2026fire}. In \emph{class-incremental learning (Class-IL)}, we split CIFAR-10 into five tasks and train ResNet-18 from scratch with DER++~\citep{buzzega2020dark}, replacing only its optimizer, to test whether direction-aware updates improve retention beyond replay; we compare against Adam, Muon, and UPGD~\citep{elsayed2024addressing} under the same 20-epoch budget. In \emph{task-incremental learning (Task-IL)}, we follow the protocol of GPM~\citep{saha2021gradient}: ten tasks from CIFAR-100, a five-layer AlexNet trained from scratch, no replay, and task identity given at test time. This is the replay-free regime of Section~\ref{subsubsec:two_capacities}, so \method{} uses its second-order memory, and we compare against GPM run from its official code. Details, extended results, and the component ablation are in Appendices~\ref{sec:more_exps_settings_appendix},~\ref{sec:more_exps_appendix}, and~\ref{sec:more_analyses_and_ablation_studies}.

\begin{table}[t]
\centering
\caption{\textbf{Domain-incremental learning on PACS.} Results averaged over three domain orders and two seeds. AP: Average Performance; AF: Average Forgetting. Best in bold, second-best underlined. Time in minutes.}
\label{tab:pacs_dil_main_results}
\small
\setlength{\tabcolsep}{3.8pt}
\renewcommand{\arraystretch}{1.10}
\resizebox{\linewidth}{!}{%
\begin{tabular}{@{}lllccccc@{}}
\toprule
\rowcolor{sectionblue}
\textbf{Regime} & \textbf{CL Method} & \textbf{Optimizer} & \textbf{Memory} 
& \textbf{Mem. (MB)} $\downarrow$ 
& \textbf{Time (m)} $\downarrow$ 
& \textbf{AP (\%)} $\uparrow$ 
& \textbf{AF (\%)} $\downarrow$ \\
\midrule

Free & None & Adam & None & $<0.1$ & \textbf{7.49} 
& $83.34 \pm 1.65$ & $15.45 \pm 3.85$ \\

Free & None & Muon & None & $<0.1$ & \underline{8.15} 
& $\underline{87.81 \pm 1.70}$ & $\underline{9.44 \pm 3.07}$ \\

Free & None & CBP & None & $<0.1$ & 12.10 
& $82.78 \pm 1.96$ & $15.94 \pm 4.18$ \\

Free & None & FIRE & None & $<0.1$ & 11.14 
& $75.94 \pm 2.34$ & $20.98 \pm 5.54$ \\

\cmidrule(lr){1-8}
\rowcolor{copehl!120}
Free & None & \textbf{\textcolor{copegreen}{FOGO}} & Centroids 
& \textcolor{copegreen}{$\underline{0.55}$} 
& \textcolor{copegreen}{9.84} 
& \textcolor{copegreen}{$\mathbf{87.89 \pm 2.94}$} 
& \textcolor{copegreen}{$\mathbf{6.77 \pm 2.31}$} \\

\midrule

Replay & ER & Adam & Buffer & 57.4 & \textbf{8.21} 
& $87.50 \pm 1.06$ & $9.81 \pm 1.62$ \\

Replay & ER & Muon & Buffer & 57.4 & \underline{8.88} 
& $\underline{89.20 \pm 1.91}$ & $\underline{7.56 \pm 0.90}$ \\

Replay & ER & CBP & Buffer & 57.4 & 13.60 
& $84.77 \pm 4.05$ & $13.01 \pm 5.70$ \\

Replay & ER & FIRE & Buffer & 57.4 & 11.52 
& $82.79 \pm 2.31$ & $13.36 \pm 0.97$ \\

\cmidrule(lr){1-8}
\rowcolor{copehl!120}
Replay & ER & \textbf{\textcolor{copegreen}{FOGO}} & Buffer + Centroids 
& \textcolor{copegreen}{$\underline{58.0}$} 
& \textcolor{copegreen}{10.44} 
& \textcolor{copegreen}{$\mathbf{90.24 \pm 3.13}$} 
& \textcolor{copegreen}{$\mathbf{4.21 \pm 0.85}$} \\

\bottomrule
\end{tabular}%
}
\end{table}

\paragraph{Results \& Ablation Study.} Table~\ref{tab:pacs_dil_main_results} shows that \method{} consistently improves retention in Domain-IL. Without replay, \method{} achieves the lowest AF, reducing forgetting by 2.67 points over Muon while matching its AP, and improving AP by 4.55 points over Adam. With ER, \method{} further improves ER+Adam by 2.74 AP points and 5.60 AF points. Notably, replay-free \method{} also \emph{surpasses ER+Adam in both AP and AF using only 0.55MB of centroid memory versus 57.4MB for replay}, with only a modest time increase. CBP and FIRE perform worse, indicating that plasticity alone is insufficient for stable domain-incremental learning.

\begin{figure*}[t]
    \centering
    \begin{subfigure}[t]{0.49\linewidth}
        \centering
        \includegraphics[width=\linewidth]{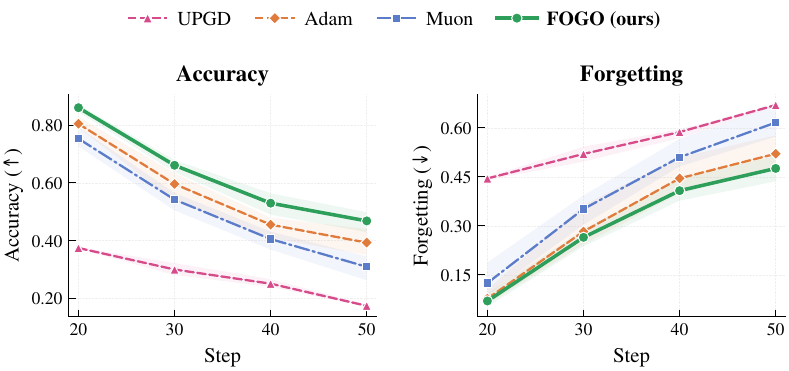}
        \caption{Comparison with Adam, Muon, and UPGD.}
        \label{fig:class_il_compare}
    \end{subfigure}
    \hfill
    \begin{subfigure}[t]{0.49\linewidth}
        \centering
        \includegraphics[width=\linewidth]{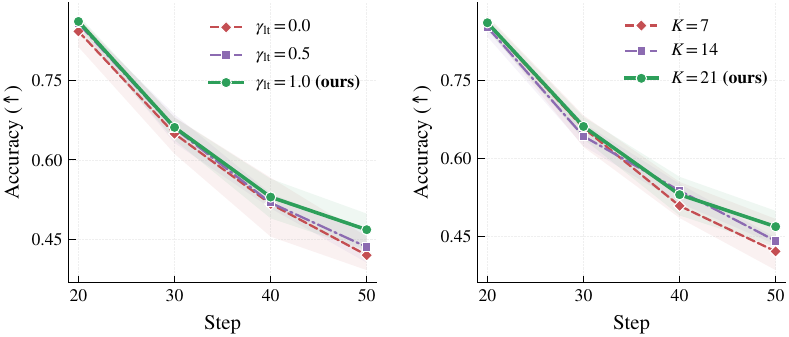}
        \caption{Ablation of long-term protection strength and memory codebook size per layer.}
        \label{fig:class_il_ablation}
    \end{subfigure}
    \caption{\textbf{Class-incremental learning on CIFAR-10.}
Left: comparison with Adam, Muon, and UPGD. Right: ablations on long-term protection strength \(\gamma_{\mathrm{lt}}\) and the number of frozen centroids per layer. Results are averaged over five random seeds.}
    \label{fig:class_il_main}
\end{figure*}

Fig.~\ref{fig:class_il_main} shows that \method{} improves Class-IL accuracy and reduces forgetting over Adam, Muon, and UPGD. Muon orthogonalizes updates but keeps no memory of which directions mattered; UPGD estimates utility from the current task, which leaves earlier class-specific directions under-protected. The ablations show that stronger long-term protection \(\gamma_{\mathrm{lt}}\) and more frozen centroids consistently improve retention. The component ladder in Appendix~\ref{sec:more_analyses_and_ablation_studies} separates the two axes of Section~\ref{sec:Method}: orthogonalized momentum under \method{}'s normalization alone reaches 40.8 AP against 31.0 for Muon, and the directional memory adds the remaining five points.

In Task-IL, where nothing revisits a past task, \method{} with second-order memory reaches \(72.05\pm1.33\) average accuracy against \(70.67\pm0.46\) for GPM over the same four seeds, at comparable memory (13.5\,MB, of which 1.8\,MB is the online codebook, versus 12.9\,MB) and without storing any sample.

\subsection{Continual Fine-tuning}

\begin{table}[t]
\centering
\caption{\textbf{Continual learning results on VQA v2.}
We compare CL-MoE and LoRAMoE with AdamW, Muon, and \method{}. AP denotes Average Performance, and SF denotes Sum Forgetting. Best results are bolded and second-best underlined.}
\label{tab:vqa_v2_results}
\small
\setlength{\tabcolsep}{4.0pt}
\renewcommand{\arraystretch}{1.10}
\resizebox{\linewidth}{!}{%
\begin{tabular}{@{}lcccccccccccc@{}}
\toprule
\multirow{2}{*}{\textbf{Method}}
& \multicolumn{10}{c}{\textbf{VQA v2 Task Types}}
& \multirow{2}{*}{$\mathbf{AP}\uparrow$}
& \multirow{2}{*}{$\mathbf{SF}\downarrow$} \\
\cmidrule(lr){2-11}
& \textbf{Rec.}
& \textbf{Loc.}
& \textbf{Jud.}
& \textbf{Com.}
& \textbf{Cou.}
& \textbf{Act.}
& \textbf{Col.}
& \textbf{Typ.}
& \textbf{Sub.}
& \textbf{Cau.}
& & \\
\midrule
CL-MoE
& 55.74 & \textbf{41.74} & 80.39 & \underline{76.77} & \underline{49.47}
& 75.41 & 73.56 & 63.17 & 61.03 & \textbf{30.41}
& 60.77 & 17.70 \\

LoRAMoE + AdamW
& \underline{55.88} & 41.45 & \underline{80.64} & 76.45 & 48.90
& 75.48 & 73.36 & 62.68 & \underline{61.97} & \underline{29.95}
& 60.68 & 18.12 \\

LoRAMoE + Muon
& \textbf{57.72} & 41.31 & 80.48 & \underline{76.77} & 48.88
& \underline{76.04} & \underline{73.73} & \underline{63.25} & 61.35 & 29.03
& \underline{60.86} & \underline{17.39} \\

\arrayrulecolor{softgold}\midrule[1.1pt]\arrayrulecolor{black}
\rowcolor{softgoldrow}
\textbf{LoRAMoE + \method{}}
& \textbf{57.72} & \underline{41.60} & \textbf{82.95} & \textbf{79.95} & \textbf{52.91}
& \textbf{77.83} & \textbf{75.56} & \textbf{63.91} & \textbf{62.72} & 29.85
& \textbf{62.50} & \textbf{7.14} \\
\bottomrule
\end{tabular}%
}
\end{table}

\paragraph{Setup.}
We evaluate continual multimodal fine-tuning on Continual VisualQA, a VQA v2~\cite{goyal2017making}-based benchmark with 200K images and 1.1M questions. Following CL-MoE~\citep{huai2025cl}, the benchmark is divided into 10 reasoning-type tasks: recognition, location, judgment, commonsense, counting, action, color, type, subcategory, and causal reasoning. We fine-tune LLaVA-1.5-7B~\citep{liu2023visual} with the LoRAMoE~\citep{dou2024loramoe} architecture and vary only the optimizer among AdamW, Muon, and \method{}, thereby isolating the effect of the update rule under the same parameter-efficient MoE setting. We report Average Performance (AP) and Sum Forgetting (SF) across all tasks. We compare against CL-MoE, since conventional continual-learning baselines were reported to perform substantially worse in~\citep{huai2025cl}. Further details are provided in Appendix~\ref{sec:more_exps_settings_appendix}.

\paragraph{Results.}
Table~\ref{tab:vqa_v2_results} shows that \method{} consistently outperforms AdamW, Muon, and the strong CL-MoE baseline, improving average accuracy by about 2 points and reducing total forgetting by roughly 10 points over the ten tasks. 
These gains suggest that \method{} better preserves important LoRA expert update directions through its compact centroid memory. 
Appendix~\ref{sec:more_analyses_and_ablation_studies} provides further analysis of training behavior, efficiency, and expert plasticity, showing that \method{} induces more balanced expert feature variance and helps mitigate expert collapse and destructive updates during continual learning.

\subsection{Language Pretraining}

\paragraph{Setup.}
We evaluate language-model pretraining from scratch on OpenWebText~\citep{Gokaslan2019OpenWeb} using GPT-2 Small (125M)~\cite{radford2019language} implemented in nanoGPT~\citep{Karpathy2022}. All methods are trained for 100K steps, corresponding to approximately 49.2B tokens, with the same architecture, data order, token budget, context length of 1024, and learning-rate schedule. We compare \method{} with Adam and Muon. Following the standard Muon hybrid setup, Muon and \method{} are applied to matrix-valued parameters, while Adam updates the remaining parameters. We report  training and validation loss over tokens to assess optimization efficiency and language-modeling performance.

\begin{figure}[t]
    \centering
    \includegraphics[width=0.48\textwidth]{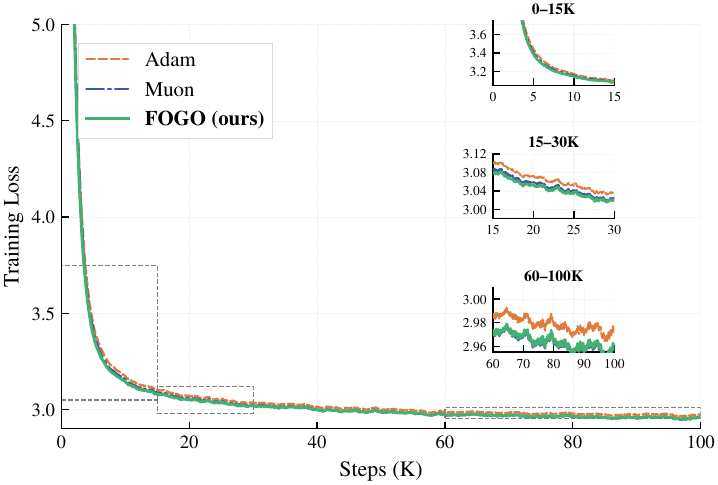}%
    \hfill
    \includegraphics[width=0.48\textwidth]{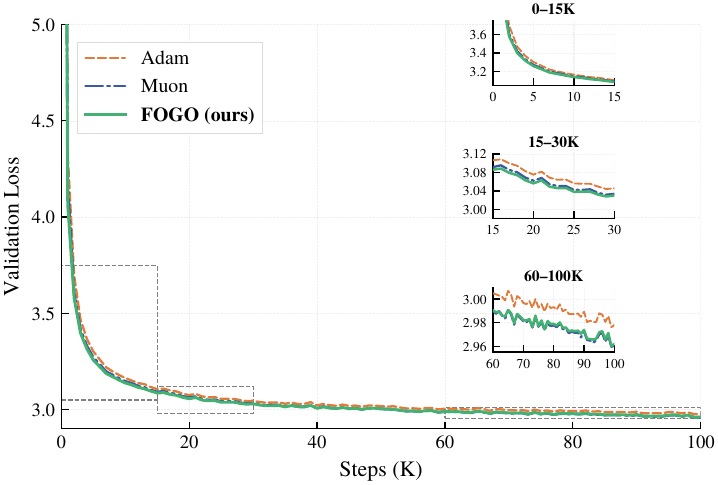}
    \caption{Training loss (left) and validation loss (right) during GPT-2 pretraining.}
    \label{fig:gpt2_pretraining}
\end{figure}

\paragraph{Results.}
Fig.~\ref{fig:gpt2_pretraining} shows that \method{} also performs well in standard unsupervised learning via the popular next-token-prediction scheme. Specifically, \method{} converges faster than both Adam and Muon during GPT-2 pretraining: the zoomed-in panels reveal a clear gap in validation loss throughout the first 30K steps, after which all three optimizers gradually approach a similar final loss. This suggests that FOGO's mechanism is beneficial not only in the continual-learning setting but also in conventional large-scale language model training.

\section{Conclusion}
\label{sec:Conclusion}

\providecommand{\rev}[1]{\textcolor{red}{#1}}%
 
 
 
 
We proposed \method{}, a scalable optimizer that unifies standard and continual learning under a single interference-resolution framework. By combining spectral orthogonalization with a compact random-projection codebook, \method{} detects and corrects gradient interference at every step without storing raw data. Across class-imbalanced learning, continual visual learning, {continual fine-tuning of LLaVA-7B}, and GPT-2 pretraining, \method{} consistently improves {retention and convergence} over Adam and Muon, and its second-order variant outperforms GPM at comparable memory, without storing any sample. Forgetting remains an open challenge, but our results suggest that {addressing} it at the optimizer level is both practical and effective. Natural extensions include {adaptive codebook sizing and centroid management, and boundary-free continual learning.}

\section*{Acknowledgements}

This work was supported by the US Army International Technology Center Pacific (ITC-IPAC) under Contract No. FA520923C0020. This research is supported by the ARC Training Centre for Whole Life Design of Carbon Neutral Infrastructure (IC230100015). The authors acknowledge the National Computational Infrastructure (NCI Australia), an NCRIS-enabled capability supported by the Australian Government, for providing computational resources used in this study. We acknowledge the Katana computational cluster, supported by Research Technology Services at UNSW Sydney.

{\small
\bibliographystyle{plainnat}
\bibliography{continual_optimiser}
}

\newpage

\appendix

\clearpage
\appendix
\begin{center}
{\Large\bfseries Appendix}\\[0.4em]
{\large FOGO: Forgetting-aware Orthogonalization Optimizer}
\end{center}

\startcontents[appendix]
\begingroup
\small
\setlength{\parskip}{0pt}
\printcontents[appendix]{}{1}{\setcounter{tocdepth}{2}}
\endgroup
\clearpage

\section{Update flow of Muon and \method{}}
\label{sec:update_flow_appendix}

\begin{figure}[t]
\centering
\includegraphics[width=\linewidth]{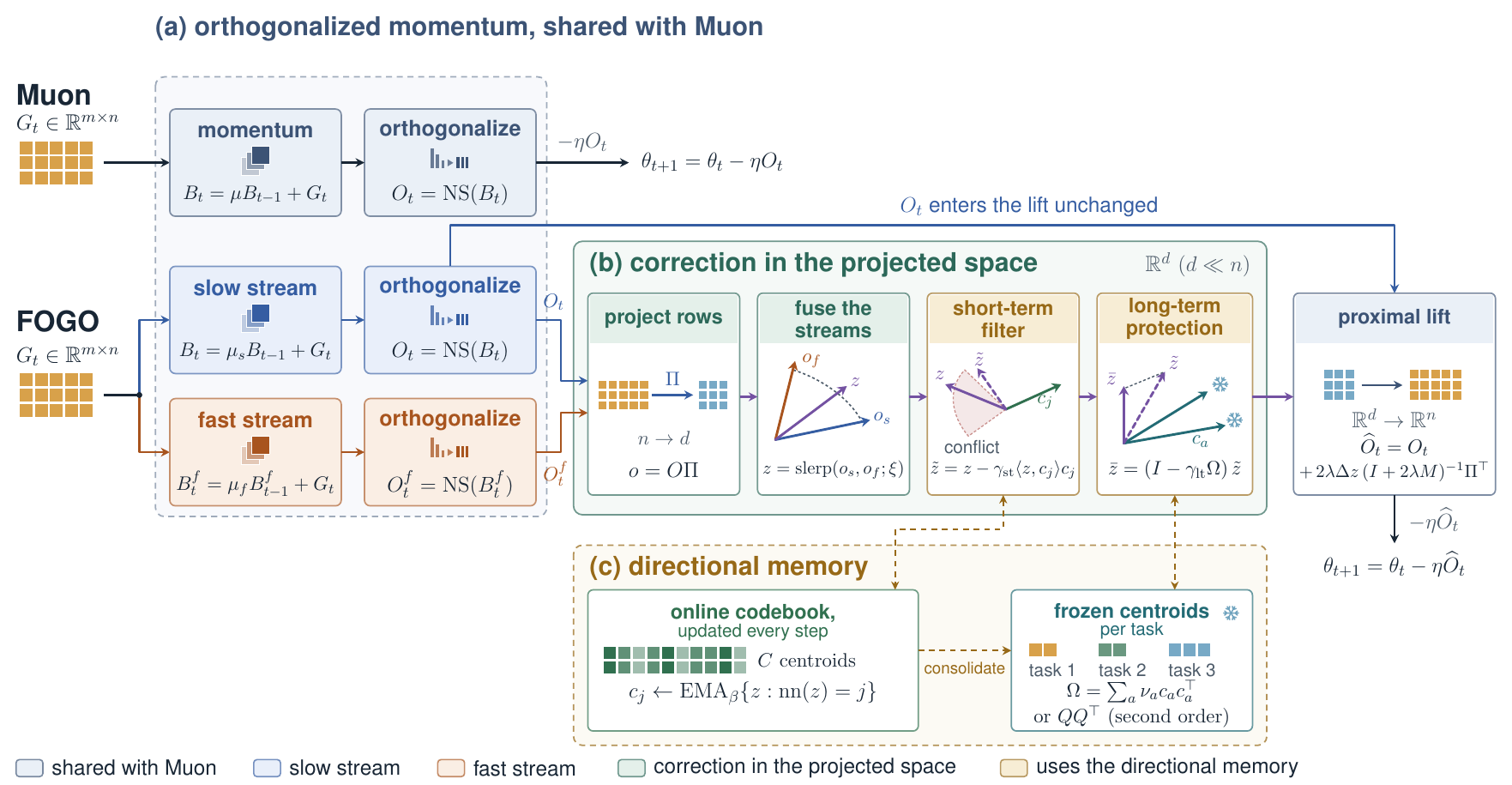}
\caption{\textbf{Update flow of Muon and \method{}.} (a) Orthogonalized momentum, shared by both. (b) \method{} projects the rows, fuses the slow and fast streams, and applies short-term filtering and long-term protection in the projected space. The proximal lift returns the correction to $\mathbb R^n$ and also takes the unchanged slow stream as input. Yellow blocks use the directional memory. (c) The online codebook is updated by EMA at each step. Each vector $z$ is assigned to its nearest centroid $c_j$, denoted by $\mathrm{nn}(z)=j$. Centroids are frozen per task. For second-order memory, the bank stores a basis $Q$ with $\Omega=QQ^\top$.}
\label{fig:update_flow}
\end{figure}

Fig.~\ref{fig:update_flow} places the two update rules side by side. Both start from the same orthogonalized momentum (panel a). \method{} then projects the rows of its slow and fast streams to $\mathbb R^{d}$, fuses them, and applies the two corrections in that space (panel b): the short-term filter deflates rows that conflict with a recurring centroid, and the long-term protection removes the component along frozen directions (Theorem~4). The proximal lift maps the corrected rows back to $\mathbb R^{n}$ with the slow stream entering unchanged, so the whole correction is a residual on Muon's update. Panel (c) shows the two memories that drive the corrections: the online codebook, updated every step, and the frozen centroids, written once per task. For the second-order memory of Section~\ref{subsubsec:two_capacities} the bank holds an orthonormal basis $Q$ and $\Omega=QQ^\top$; the update flow is otherwise identical.

\section{Extended Background}
\label{sec:extended_background}
 
\paragraph{Orthogonalization.}
Let $B_t \in \mathbb{R}^{m \times n}$ with compact SVD $B_t = U_t \Sigma_t V_t^\top$, where $\Sigma_t = \mathrm{diag}(\sigma_1, \dots, \sigma_r)$ and $\sigma_1 \ge \cdots \ge \sigma_r > 0$. The orthogonalized update $\mathcal{O}(B_t) := U_t V_t^\top$ discards $\Sigma_t$ while retaining the left and right singular directions. Intuitively, this equalizes the influence of all singular modes: rather than allowing a few high-energy directions to dominate the step, orthogonalization reweights the update toward under-represented directions. \citet{bernstein2024old} formalize this by showing that Muon and related optimizers perform steepest descent under a spectral-norm constraint.
 

\paragraph{Newton--Schulz iterations.}
Computing the polar factor $\mathcal{O}(B_t)$ exactly requires an SVD, which is too expensive to run at every step. Newton--Schulz iterations avoid this: starting from $X_0 = B_t/\|B_t\|_F$, each iteration applies a fixed polynomial to $X_k$ that moves all singular values closer to $1$ without changing the singular vectors. After a few iterations, $X_T$ is close to $\mathcal{O}(B_t)$. We follow Muon~\citep{jordan2024muon} and use the quintic polynomial $p(\sigma)=a\sigma+b\sigma^3+c\sigma^5$ with $(a,b,c)=(3.4445,-4.7750,2.0315)$ and $T=5$.
 
\paragraph{G-Vendi Score.}
G-Vendi Score~\citep{jung2025prismatic} extends the Vendi score~\citep{friedman2023the} from sample embeddings to the gradient space of deep networks. For each sample $i$, the per-sample gradient $G_i = \nabla_\theta \mathcal{L}(\theta; x_i, y_i) \in \mathbb{R}^{|\theta|}$ is projected via $G_i^{\mathrm{proj}} = \Pi^\top G_i$. The G-Vendi Score is computed from the Gram matrix $K \in \mathbb{R}^{N \times N}$ of projected gradients, where $K_{ij}$ denotes the normalized similarity between $G_i^{\mathrm{proj}}$ and $G_j^{\mathrm{proj}}$. This provides a model-centric embedding whose clustering structure reveals representative update directions and the geometry of the gradient distribution.

\section{Theoretical Results}
\label{sect:theoretical_results_appendix}

\subsection{Proof of Theorem 1}

Recall that $\mathcal P_t^{(r)}(X)
=
U_t^{(r)}U_t^{(r)\top}\,X\,V_t^{(r)}V_t^{(r)\top}$ and 
$\mathcal P_t^{(r)\perp}(X)=X-\mathcal P_t^{(r)}(X),$
so both \(\mathcal P_t^{(r)}\) and \(\mathcal P_t^{(r)\perp}\) are linear operators on \(\mathbb R^{m\times n}\). Under the momentum update $B_t=\mu B_{t-1}+G_t,$ we have
\begin{equation}
R_t
=
\|\mathcal P_t^{(r)\perp}(B_t)\|_F
=
\|\mathcal P_t^{(r)\perp}(\mu B_{t-1}+G_t)\|_F .
\end{equation}
By linearity of \(\mathcal P_t^{(r)\perp}\) and the triangle inequality,
\begin{equation}
R_t
\le
\mu \|\mathcal P_t^{(r)\perp}(B_{t-1})\|_F
+
\|\mathcal P_t^{(r)\perp}(G_t)\|_F
=
\mu \|\mathcal P_t^{(r)\perp}(B_{t-1})\|_F + \varepsilon_t.
\end{equation}

It remains to bound \(\|\mathcal P_t^{(r)\perp}(B_{t-1})\|_F\). By definition,
\begin{equation}
\mathcal P_t^{(r)\perp}(B_{t-1})
=
(I-\mathcal P_t^{(r)})(B_{t-1}).
\end{equation}
Adding and subtracting \(\mathcal P_{t-1}^{(r)}(B_{t-1})\) gives
\begin{equation}
(I-\mathcal P_t^{(r)})(B_{t-1})
=
(I-\mathcal P_{t-1}^{(r)})(B_{t-1})
+
(\mathcal P_{t-1}^{(r)}-\mathcal P_t^{(r)})(B_{t-1}).
\end{equation}
Taking Frobenius norms and applying the triangle inequality yields
\begin{equation}
\|\mathcal P_t^{(r)\perp}(B_{t-1})\|_F
\le
\|(I-\mathcal P_{t-1}^{(r)})(B_{t-1})\|_F
+
\|(\mathcal P_{t-1}^{(r)}-\mathcal P_t^{(r)})(B_{t-1})\|_F .
\end{equation}
Since
\[
\|(I-\mathcal P_{t-1}^{(r)})(B_{t-1})\|_F = R_{t-1},
\]
we obtain
\begin{equation}
\|\mathcal P_t^{(r)\perp}(B_{t-1})\|_F
\le
R_{t-1}
+
\|(\mathcal P_{t-1}^{(r)}-\mathcal P_t^{(r)})(B_{t-1})\|_F .
\end{equation}
By the slow-drift assumption,
\[
\|\mathcal P_t^{(r)}-\mathcal P_{t-1}^{(r)}\|_{F\to F}\le \delta_t,
\]
hence
\begin{equation}
\|(\mathcal P_{t-1}^{(r)}-\mathcal P_t^{(r)})(B_{t-1})\|_F
\le
\delta_t \|B_{t-1}\|_F .
\end{equation}
Therefore,
\begin{equation}
\|\mathcal P_t^{(r)\perp}(B_{t-1})\|_F
\le
R_{t-1}+\delta_t \|B_{t-1}\|_F .
\end{equation}
Substituting this bound into the earlier inequality gives
\begin{equation}
R_t
\le
\mu R_{t-1}
+
\mu \delta_t \|B_{t-1}\|_F
+
\varepsilon_t,
\end{equation}
which proves the claim. \hfill $\square$

\subsection{Proof of Corollary 5}

\paragraph{Corollary 5 (Unrolled suppression bound).}
Under the assumptions of Theorem~1, let
\[
a_t := \mu \delta_t \|B_{t-1}\|_F + \varepsilon_t.
\]
Then the off-subspace residual satisfies
\begin{equation}
R_t
\le
\mu^t R_0
+
\sum_{k=1}^t \mu^{\,t-k} a_k.
\end{equation}
In particular, if \(a_t \le \bar a\) for all \(t\), then
\begin{equation}
R_t
\le
\mu^t R_0 + \bar a \frac{1-\mu^t}{1-\mu},
\end{equation}
and hence
\begin{equation}
\limsup_{t\to\infty} R_t \le \frac{\bar a}{1-\mu}.
\end{equation}
Therefore, the off-subspace residual contracts geometrically up to a steady-state floor of order \(\bar a/(1-\mu)\). In particular, if \(a_t \to 0\), then \(R_t \to 0\).

\paragraph{Proof of Corollary 5.}
From Theorem~1,
\begin{equation}
R_t
\le
\mu R_{t-1}
+
\mu \delta_t \|B_{t-1}\|_F
+
\varepsilon_t
=
\mu R_{t-1}+a_t.
\end{equation}
Iterating this recursion yields
\begin{equation}
R_t
\le
\mu^t R_0+\sum_{k=1}^t \mu^{\,t-k} a_k.
\end{equation}
This proves the first claim. If \(a_t\le \bar a\) for all \(t\), then
\begin{equation}
R_t
\le
\mu^t R_0+\bar a\sum_{k=1}^t \mu^{\,t-k}
=
\mu^t R_0+\bar a\frac{1-\mu^t}{1-\mu}.
\end{equation}
Since \(0\le \mu<1\), taking \(t\to\infty\) gives
\begin{equation}
\limsup_{t\to\infty} R_t \le \frac{\bar a}{1-\mu}.
\end{equation}
Thus, \(R_t\) contracts geometrically up to a steady-state floor of order \(\bar a/(1-\mu)\).

Finally, if \(a_t\to 0\), then for any \(\epsilon>0\) there exists \(T\) such that \(a_k\le \epsilon\) for all \(k\ge T\). Splitting the sum at \(T\),
\begin{equation}
\sum_{k=1}^t \mu^{\,t-k} a_k
=
\sum_{k=1}^{T-1} \mu^{\,t-k} a_k
+
\sum_{k=T}^t \mu^{\,t-k} a_k.
\end{equation}
The first term tends to \(0\) as \(t\to\infty\), because \(T\) is fixed and thus it contains only finitely many terms \(\mu^{\,t-k}a_k\), each converging to \(0\). For the second term,
\begin{equation}
\sum_{k=T}^t \mu^{\,t-k} a_k
\le
\epsilon \sum_{k=T}^t \mu^{\,t-k}
\le
\frac{\epsilon}{1-\mu}.
\end{equation}
Hence
\[
\limsup_{t\to\infty} \sum_{k=1}^t \mu^{\,t-k} a_k
\le
\frac{\epsilon}{1-\mu}.
\]
Since this is true for every \(\epsilon>0\), the limsup is at most \(0\). As the sum is nonnegative for all \(t\), its liminf is at least \(0\). Therefore,
\[
\sum_{k=1}^t \mu^{\,t-k} a_k \to 0.
\]
Together with \(\mu^tR_0\to 0\), this implies \(R_t\to 0\). \hfill $\square$

This shows that off-subspace directions are suppressed unless they are continually replenished by new gradient energy or sustained by persistent subspace drift.

\subsection{Proof of Corollary 2 and the Full-Batch Remark}
\label{subsec:proof_cor2}

Since \(\mathcal L_S\) is \(\beta\)-smooth, for all \(\theta,\theta'\),
\[
\mathcal L_S(\theta')
\ge
\mathcal L_S(\theta)
+
\langle\nabla\mathcal L_S(\theta),\theta'-\theta\rangle
-
\frac{\beta}{2}\|\theta'-\theta\|_2^2 .
\]
Setting \(\theta=\theta_t\) and \(\theta'=\theta_{t+1}=\theta_t-\eta u_t\) gives
\[
\mathcal L_S(\theta_{t+1})
\ge
\mathcal L_S(\theta_t)
-\eta\langle\nabla\mathcal L_S(\theta_t),u_t\rangle
-\frac{\beta\eta^2}{2}\|u_t\|_2^2 .
\]
The right-hand side exceeds \(\mathcal L_S(\theta_t)\) exactly when
\(-\eta\langle\nabla\mathcal L_S(\theta_t),u_t\rangle-\frac{\beta\eta^2}{2}\|u_t\|_2^2>0\),
which for \(\eta>0\) is the condition
\(\langle\nabla\mathcal L_S(\theta_t),u_t\rangle<-\frac{\beta\eta}{2}\|u_t\|_2^2\)
stated in Corollary~2. \hfill $\square$

\paragraph{Full-batch descent.}
Write the full objective as \(\mathcal L=p_D\mathcal L_D+p_R\mathcal L_R\) with \(g_D=\nabla\mathcal L_D(\theta_t)\) and \(g_R=\nabla\mathcal L_R(\theta_t)\), so the exact full-batch direction is \(u_t=\nabla\mathcal L(\theta_t)=p_Dg_D+p_Rg_R\). Applying Corollary~2 with \(S\) the rare group, \(\nabla\mathcal L_S=g_R\) and \(\langle g_R,u_t\rangle=p_D\langle g_R,g_D\rangle+p_R\|g_R\|_2^2\), so the rare-group loss strictly increases whenever
\[
p_D\langle g_R,g_D\rangle+p_R\|g_R\|_2^2
<
-\frac{\beta\eta}{2}\,\|p_Dg_D+p_Rg_R\|_2^2 .
\]
Dividing by \(p_D>0\) and letting \(\eta\to0\) gives the condition \(\langle g_R,g_D\rangle<-\frac{p_R}{p_D}\|g_R\|_2^2\) quoted in the main text, whose threshold shrinks toward zero as \(p_R/p_D\) decreases. No property of mini-batching enters the argument: \(g_D\) and \(g_R\) are determined by the objective at \(\theta_t\), and batching only changes how noisily they are estimated.

\subsection{Proof of Corollary 3}
\label{subsec:proof_cor3}

Taking the Frobenius inner product of \(B_t=\mu B_{t-1}+G_t\) with \(H\) gives the exact scalar recursion
\[
b_t=\mu b_{t-1}+\langle G_t,H\rangle_F .
\]
Conditioning on \(\mathcal F_{t-1}\), under which \(b_{t-1}\) is fixed, the assumption gives
\[
\mu b_{t-1}+\kappa_-
\le
\mathbb E[b_t\mid\mathcal F_{t-1}]
\le
\mu b_{t-1}+\kappa_+ .
\]
For \(t>s\), let \(m_t:=\mathbb E[b_t\mid\mathcal F_s]\). Taking \(\mathbb E[\,\cdot\mid\mathcal F_s]\) and using the tower property,
\(\mu m_{t-1}+\kappa_-\le m_t\le\mu m_{t-1}+\kappa_+\), with \(m_s=b_s\). Unrolling from \(s\) to \(t\),
\[
\mu^{t-s}b_s+\kappa_-\sum_{j=0}^{t-s-1}\mu^j
\le
m_t
\le
\mu^{t-s}b_s+\kappa_+\sum_{j=0}^{t-s-1}\mu^j ,
\]
and \(\sum_{j=0}^{t-s-1}\mu^j=(1-\mu^{t-s})/(1-\mu)\). \hfill $\square$

\paragraph{The three regimes.}
Under imbalance, suppose each step contains the rare group with probability \(p_R\), a step that contains it contributes \(\langle G_t,H\rangle_F\ge\alpha>0\), and any other step contributes at least \(-\gamma\). Then
\(\mathbb E[\langle G_t,H\rangle_F\mid\mathcal F_{t-1}]\ge p_R\alpha-(1-p_R)\gamma=:\kappa_-\),
and the lower bound gives \(\liminf_{t\to\infty}\mathbb E[b_t\mid\mathcal F_s]\ge\kappa_-/(1-\mu)\), a level that scales with \(p_R\) and is non-positive once \(p_R\alpha\le(1-p_R)\gamma\). Under reshuffled balanced training the same bound holds with \(\kappa_->0\). Under task shift with \(\kappa_+\le0\), the upper bound gives \(\mathbb E[b_t\mid\mathcal F_s]\le\mu^{t-s}b_s\), which decays geometrically.

\subsection{Spectral Equalization by Orthogonalization}
\label{subsec:appendix_orthogonalization_spectral}

\paragraph{Proposition 6 (Orthogonalization as Spectral equalization).}
\label{prop:spectral_equalization}
Let \(B=U\Sigma V^\top\in\mathbb R^{m\times n}\) have rank \(k\), with singular values
\(\sigma_1\ge\cdots\ge\sigma_k>0\), and let \(O=UV^\top\) be its orthogonalized update. For each singular mode \(E_i=u_i v_i^\top\),
\[
\langle B,E_i\rangle_F=\sigma_i,
\qquad
\langle O,E_i\rangle_F=1 .
\]
Hence, for any \(i,j\le k\),
\[
\frac{\langle B,E_i\rangle_F}{\langle B,E_j\rangle_F}
=
\frac{\sigma_i}{\sigma_j},
\qquad
\frac{\langle O,E_i\rangle_F}{\langle O,E_j\rangle_F}
=
1 .
\]
Thus, orthogonalization preserves the singular directions of \(B\) while equalising their coefficients. In particular, a weak mode \(E_i\) receives a relative amplification factor \(\sigma_j/\sigma_i\) compared with any stronger mode \(E_j\).

\paragraph{Proof of Proposition 6}
Since
\[
B=\sum_{i=1}^k \sigma_i u_i v_i^\top,
\qquad
O=UV^\top=\sum_{i=1}^k u_i v_i^\top,
\]
each singular mode \(E_i=u_i v_i^\top\) appears in \(B\) with coefficient \(\sigma_i\), and in \(O\) with coefficient \(1\). Moreover, the rank-one matrices \(\{E_i\}_{i=1}^k\) are orthonormal under the Frobenius inner product:
\[
\langle E_i,E_j\rangle_F
=
\mathrm{tr}(v_i u_i^\top u_j v_j^\top)
=
(u_i^\top u_j)(v_i^\top v_j)
=
\delta\{i=j\}.
\]
Therefore,
\[
\langle B,E_i\rangle_F=\sigma_i,
\qquad
\langle O,E_i\rangle_F=1,
\]
The ratio statements follow directly. Intuitively, orthogonalization equalizes all retained singular modes, allowing weak but useful directions to remain visible rather than being overwhelmed by dominant modes.

\subsection{Gradient Geometry Preservation under Random Projection}
\label{subsec:gradient_geometry_random_projection}

We justify operating on projected rows rather than full rows. \method{} projects
each row of a layer's update, so let
\(\mathcal G=\{g_i\}_{i=1}^N\subset\mathbb R^{n}\) be the finite set of row
vectors and centroids compared at a step, and let
\(\Pi\in\mathbb R^{n\times d}\) be a Rademacher random projection with
entries scaled by \(1/\sqrt d\). By the Johnson--Lindenstrauss lemma, for any
\(\epsilon\in(0,1)\), if \(d=\mathcal O(\epsilon^{-2}\log N)\), then with high
probability, for all \(i,j\),
\[
(1-\epsilon)\|g_i-g_j\|_2^2
\le
\|\Pi^\top g_i-\Pi^\top g_j\|_2^2
\le
(1+\epsilon)\|g_i-g_j\|_2^2 .
\]
Thus, projected rows preserve pairwise geometry up to small multiplicative
distortion. This supports clustering and conflict detection in the projected
space, while reducing the dimension of every stored or compared vector from
\(n\) to \(d\). The required \(d\) depends logarithmically on the number of
compared vectors \(N\), not on the row width \(n\); in practice, dimensions
such as \(d=128\) to \(512\) already provide a compact surrogate for online
directional memory. Cosine similarities, which all codebook operations use, are
invariant to the \(1/\sqrt d\) scale, so the guarantee applies whether or not
the scale is applied in the implementation.

\paragraph{Proof Sketch.}
The result follows directly from the Johnson--Lindenstrauss lemma~\citep{johnson1984extensions} applied to the
finite set \(\mathcal G\). A Rademacher matrix with \(1/\sqrt d\) scaling is a
sub-Gaussian Johnson--Lindenstrauss transform. Hence, for any fixed pair
\((g_i,g_j)\), the projected squared distance
\(\|\Pi^\top(g_i-g_j)\|_2^2\) concentrates around
\(\|g_i-g_j\|_2^2\) within a factor \(1\pm\epsilon\). A union bound over all
\(N(N-1)/2\) distinct pairs gives the stated guarantee when
\(d=\mathcal O(\epsilon^{-2}\log N)\).

\subsection{Derivation of the Proximal Lifting Solution}
\label{subsec:proximal_lifting_derivation}

We derive the closed-form solution used to lift projected corrections back to
the original update space. Recall that \(\Pi\in\mathbb R^{n\times d}\) is the
fixed random projection, \(P_{t,i}^{(s)}=O_{t,i}^{(s)}\Pi\in\mathbb R^d\) is
the projected slow-stream update, and \(\bar z_{t,i}\in\mathbb R^d\) is the
memory-protected direction after short- and long-term filtering. The projected
correction is
\[
\Delta z_{t,i}=\bar z_{t,i}-P_{t,i}^{(s)}\in\mathbb R^d.
\]
Since \(\Pi\) is not invertible, we find the smallest correction
\(\delta_{t,i}\in\mathbb R^n\) whose projection matches \(\Delta z_{t,i}\) in a
least-squares sense:
\[
\delta_{t,i}^{\mathrm{prox}}
=
\arg\min_{\delta\in\mathbb R^n}
\left\{
\frac12\|\delta\|_2^2
+
\lambda\|\delta^\top\Pi-\Delta z_{t,i}\|_2^2
\right\}.
\]
Let
\[
f(\delta)
=
\frac12\|\delta\|_2^2
+
\lambda\|\Pi^\top\delta-\Delta z_{t,i}\|_2^2 .
\]
Taking the gradient gives
\[
\nabla_\delta f
=
\delta
+
2\lambda\Pi(\Pi^\top\delta-\Delta z_{t,i}).
\]
Setting \(\nabla_\delta f=0\) yields
\[
(I+2\lambda\Pi\Pi^\top)\delta
=
2\lambda\Pi\Delta z_{t,i},
\]
and hence the direct high-dimensional solution is
\[
\delta_{t,i}^{\mathrm{prox}}
=
(I+2\lambda\Pi\Pi^\top)^{-1}
(2\lambda\Pi\Delta z_{t,i}).
\]
Using the identity
\[
(I+2\lambda\Pi\Pi^\top)^{-1}\Pi
=
\Pi(I+2\lambda\Pi^\top\Pi)^{-1},
\]
we obtain the equivalent low-dimensional form
\[
\delta_{t,i}^{\mathrm{prox}}
=
\Pi(I+2\lambda\Pi^\top\Pi)^{-1}(2\lambda\Delta z_{t,i}).
\]
Therefore, with \(M=\Pi^\top\Pi\), we compute
\[
u_{t,i}^{\mathrm{prox}}
=
(I+2\lambda M)^{-1}(2\lambda\Delta z_{t,i}),
\qquad
\delta_{t,i}^{\mathrm{prox}}
=
\Pi u_{t,i}^{\mathrm{prox}}.
\]
The objective is strongly convex for \(\lambda>0\), so the minimizer is unique.

\subsection{Proof of Theorem 4}

For every $a\in\mathcal I$ the rank-one matrix $c_ac_a^\top$ is symmetric
positive semidefinite, and $\nu_a\ge0$, so
$\Omega=\sum_{a\in\mathcal I}\nu_a c_ac_a^\top$ is symmetric positive
semidefinite. Write its eigendecomposition as
\[
\Omega=W\Lambda W^\top,
\qquad
\Lambda=\mathrm{diag}(\lambda_1,\ldots,\lambda_d),
\qquad
\lambda_j\ge0,
\]
with $W$ orthogonal, and let $y=W^\top x$. From $x^{+}=(I-\gamma\Omega)x$ we get
\[
W^\top x^{+}=(I-\gamma\Lambda)\,y,
\]
so in the eigenbasis of $\Omega$ the $j$-th coordinate of the update is
multiplied by $1-\gamma\lambda_j$.

\emph{Contraction.} The energy measured by $\Omega$ after protection is
\[
\|\Omega^{1/2}x^{+}\|_2^2
=\|\Lambda^{1/2}(I-\gamma\Lambda)y\|_2^2
=\sum_{j=1}^{d}\lambda_j\,(1-\gamma\lambda_j)^2\,y_j^2 .
\]
If $0\le\gamma\le1/\lambda_{\max}(\Omega)$, then $0\le1-\gamma\lambda_j\le1$
for every $j$, hence $\lambda_j(1-\gamma\lambda_j)^2y_j^2\le\lambda_jy_j^2$.
Summing over $j$,
\[
\|\Omega^{1/2}x^{+}\|_2^2\le\sum_{j=1}^{d}\lambda_jy_j^2=\|\Omega^{1/2}x\|_2^2 ,
\]
and taking square roots gives the first claim.

\emph{Free directions.} If $\Omega x=0$, then
$x^{+}=x-\gamma\Omega x=x$, which is the second claim.

\emph{Second-order memory.} With $Q=[q_1,\ldots,q_r]$ orthonormal,
$\Omega=QQ^\top=\sum_{j=1}^{r}q_jq_j^\top$ has the form of the theorem with
$\nu_j=1$, in $\mathbb R^{n}$ instead of $\mathbb R^{d}$. It is the orthogonal
projector onto $\mathrm{span}(Q)$, so $\lambda_{\max}(\Omega)=1$ and the
condition is $\gamma\le1$. With $\gamma=1$,
\[
x^{+}=(I-QQ^\top)\,x ,
\]
which removes the component of $x$ in $\mathrm{span}(Q)$ and leaves the
orthogonal component unchanged. This is the hard projection used in the
Task-IL experiments.

\begin{algorithm}[t]
\caption{FOGO: Forgetting-aware Orthogonalization Optimizer}
\label{alg:FOGO_compact}
\begin{algorithmic}[1]
\Require task sequence $\{\mathcal T_r\}_{r=1}^T$, learning rate $\eta$, weight decay $\mathrm{wd}$,
projection $\Pi$, hyperparameters, and $\varepsilon>0$
\State Initialize long-term memory $\mathcal M_{<1}^{\mathrm{lt}}\gets\emptyset$ and $M\gets\Pi^\top\Pi$
\For{task $r=1,\dots,T$}
    \State Initialize $B_0^{(s)},B_0^{(f)}\gets0$, active codebook $\mathcal C_{r,0}$, and conflict EMA $\bar r_0\gets0$
    \For{optimization step $t=1,\dots,U_r$}
        \State $G_t\gets\nabla_\theta\mathcal L_{\mathcal T_r}(\theta)$

        \Statex
        \State \textbf{\slowblk{Slow--fast streams:}}
        Update $B_t^{(s)},B_t^{(f)}$; orthogonalize them into
        $O_t^{(s)},O_t^{(f)}$; project to
        $P_t^{(s)}=O_t^{(s)}\Pi$ and $P_t^{(f)}=O_t^{(f)}\Pi$

        \Statex
        \State \textbf{\slerpblk{Geodesic fusion:}}
        Fuse each projected row pair $(P_{t,i}^{(s)},P_{t,i}^{(f)})$ into
        $z'_{t,i}$ by \textsc{Slerp}; set
        $z_{t,i}=\|P_{t,i}^{(s)}\|_2 z'_{t,i}$ and
        $q_{t,i}=z_{t,i}/(\|z_{t,i}\|_2+\varepsilon)$

        \Statex
        \State \textbf{\memblk{Online memory read and accumulation:}}
        Compare $\{q_{t,i}\}_i$ with the pre-update codebook
        $\mathcal C_{r,t-1}$; assign each query to its nearest centroid and
        accumulate $(m_{t,j},s_{t,j})$

        \Statex
        \State \textbf{\corrblk{Short-term filtering:}}
        Split normalized centroids from $\mathcal C_{r,t-1}$ into safe,
        over-aligned, and conflicting sets; compute
        $z_{t,i}^{\mathrm{hi}},z_{t,i}^{\mathrm{conf}}$; update
        $\gamma_{\mathrm{st},t}$ from $\bar r_t$
        \Statex \hspace{\algorithmicindent}
        Set
        $\tilde z_{t,i}
        =
        z_{t,i}
        -
        \gamma_{\mathrm{hi}}z_{t,i}^{\mathrm{hi}}
        -
        \gamma_{\mathrm{st},t}z_{t,i}^{\mathrm{conf}}$

        \Statex
        \State \textbf{\forgetblk{Long-term protection:}}
        Let $\mathcal A_t^{\mathrm{lt}}\subseteq\mathcal M_{<r}^{\mathrm{lt}}$
        be the available frozen centroids from previous tasks; form
        $\tilde q_{t,i}=\tilde z_{t,i}/(\|\tilde z_{t,i}\|_2+\varepsilon)$
        \Statex \hspace{\algorithmicindent}
        Activate
        $\mathcal B_{t,i}^{\mathrm{lt}}
        =
        \{(\mathcal T_\tau,k)\in\mathcal A_t^{\mathrm{lt}}:
        |\langle\tilde q_{t,i},\bar c_{\mathcal T_\tau,k}\rangle|>\epsilon_{\mathrm{lt}}\}$,
        where $\tau<r$
        \Statex \hspace{\algorithmicindent}
        Compute $z_{t,i}^{\mathrm{lt}}$ from $\mathcal B_{t,i}^{\mathrm{lt}}$ and set
        $\bar z_{t,i}=\tilde z_{t,i}-\gamma_{\mathrm{lt}}z_{t,i}^{\mathrm{lt}}$

        \Statex
        \State \textbf{\proxblk{Proximal lifting:}}
        Lift $\Delta z_{t,i}=\bar z_{t,i}-P_{t,i}^{(s)}$ by
        $u_{t,i}^{\mathrm{prox}}=(I+2\lambda M)^{-1}(2\lambda\Delta z_{t,i})$
        \Statex \hspace{\algorithmicindent}
        Set $\hat O_{t,i}=O_{t,i}^{(s)}+\Pi u_{t,i}^{\mathrm{prox}}$

        \Statex
        \State \textbf{\proxblk{Parameter update:}}
        Stack $\{\hat O_{t,i}\}_i$ into $\hat O_t$;
        set $\gamma_t=\sigma\sqrt{mn}/(\|\hat O_t\|_F+\varepsilon)$
        \Statex \hspace{\algorithmicindent}
        Update $\theta\gets(1-\eta\,\mathrm{wd})\theta-\eta\gamma_t\hat O_t$

        \Statex
        \State \textbf{\memblk{Codebook flush:}}
        Update EMA statistics $(S_{t,j},N_{t,j})$ using accumulated
        $(m_{t,j},s_{t,j})$; normalize, decorrelate, and re-seed centroids to
        obtain $\mathcal C_{r,t}$
    \EndFor

    \Statex
    \State \textbf{\forgetblk{Consolidation:}}
    Store top-$K_r$ centroids from $\mathcal C_{r,U_r}$ as
    $\mathcal M_r^{\mathrm{lt}}
    =
    \{(\bar c_{\mathcal T_r,k},\nu_{\mathcal T_r,k})\}_{k=1}^{K_r}$
    \Statex \hspace{\algorithmicindent}
    Update $\mathcal M_{<r+1}^{\mathrm{lt}}\gets
    \mathcal M_{<r}^{\mathrm{lt}}\cup\mathcal M_r^{\mathrm{lt}}$
\EndFor
\State \Return $\theta$
\end{algorithmic}
\end{algorithm}

\section{Detailed Algorithm}

We provide the full \method{} algorithm in Algorithm~\ref{alg:FOGO_compact},
showing how it addresses both short-term and long-term forgetting in standard and continual learning settings.

\subsection{Details of Slow--Fast Fusion}
\label{subsec:detail_fusion}

We expand the row-wise \(\textsc{Slerp}\) operation used in the main text. For
\(P_{t,i}^{(s)},P_{t,i}^{(f)}\in\mathbb R^d\), define the normalized slow and
fast directions
\[
\bar p_{t,i}^{(s)}
=
\frac{P_{t,i}^{(s)}}{\|P_{t,i}^{(s)}\|_2+\varepsilon},
\qquad
\bar p_{t,i}^{(f)}
=
\frac{P_{t,i}^{(f)}}{\|P_{t,i}^{(f)}\|_2+\varepsilon}.
\]
Let
\[
\alpha_{t,i}^{\mathrm{sf}}
=
\langle \bar p_{t,i}^{(f)},\bar p_{t,i}^{(s)}\rangle,
\qquad
\theta_{t,i}
=
\arccos\!\left(
\mathrm{clip}(\alpha_{t,i}^{\mathrm{sf}},-1+\varepsilon,1-\varepsilon)
\right).
\]
The tangent direction from the slow stream toward the fast stream is
\[
v_{t,i}
=
\frac{
\bar p_{t,i}^{(f)}
-
\alpha_{t,i}^{\mathrm{sf}}\bar p_{t,i}^{(s)}
}{
\left\|
\bar p_{t,i}^{(f)}
-
\alpha_{t,i}^{\mathrm{sf}}\bar p_{t,i}^{(s)}
\right\|_2+\varepsilon
}.
\]
Therefore,
\[
z'_{t,i}
=
\cos(\xi\theta_{t,i})\bar p_{t,i}^{(s)}
+
\sin(\xi\theta_{t,i})v_{t,i}.
\]
The scaled fused update is then obtained as in the main text,
\[
z_{t,i}
=
\|P_{t,i}^{(s)}\|_2 z'_{t,i}.
\]
This keeps the slow-stream magnitude while rotating its direction toward the
fast stream by a fraction \(\xi\) of the slow--fast angle.

\subsection{Details of Online Directional Memory}
\label{subsec:detail_online_directional_memory}

We provide the codebook update details omitted from the main text. After assigning
queries to the pre-update codebook, we collect for each centroid \(j\) the
assigned mass and directional sum:
\[
m_{t,j}
=
\sum_{i=1}^m
\mathbf 1[a_{t,i}^{\mathrm{mem}}=j],
\qquad
s_{t,j}
=
\sum_{i=1}^m
\mathbf 1[a_{t,i}^{\mathrm{mem}}=j]q_{t,i}.
\]
The codebook maintains EMA sufficient statistics,
\[
S_{t,j}
=
\beta_{\mathrm{cb}}S_{t-1,j}
+
(1-\beta_{\mathrm{cb}})s_{t,j},
\qquad
N_{t,j}
=
\beta_{\mathrm{cb}}N_{t-1,j}
+
(1-\beta_{\mathrm{cb}})m_{t,j},
\]
from which we form a provisional centroid
\[
\tilde c_{t,j}
=
\frac{S_{t,j}}{N_{t,j}+\varepsilon}.
\]

\paragraph{Codebook decorrelation and re-seeding.}
To avoid redundant memory slots, we mildly decorrelate each provisional centroid
from its most correlated pre-update centroids. Let
\[
\hat c_{t,j}
=
\frac{\tilde c_{t,j}}{\|\tilde c_{t,j}\|_2+\varepsilon},
\qquad
\mathcal R_{t,j}
=
\textsc{TopR}_{\ell\neq j}
\left|
\langle \hat c_{t,j},\bar c_{t-1,\ell}\rangle
\right|
\]
denote the \(R\) previous centroids most correlated with \(\hat c_{t,j}\). The
redundant component is
\[
h_{t,j}^{\mathrm{red}}
=
\sum_{\ell\in\mathcal R_{t,j}}
\langle \hat c_{t,j},\bar c_{t-1,\ell}\rangle
\bar c_{t-1,\ell},
\]
and the updated centroid is
\[
c_{t,j}
=
\frac{\tilde c_{t,j}-\lambda_c h_{t,j}^{\mathrm{red}}}
{\|\tilde c_{t,j}-\lambda_c h_{t,j}^{\mathrm{red}}\|_2+\varepsilon}.
\]
The coefficient \(\lambda_c\) is kept small so that centroids remain faithful to
their assigned directions while avoiding redundant slots. Centroids with
persistently small usage \(N_{t,j}\) are periodically re-seeded from the current
queries \(\{q_{t,i}\}_{i=1}^m\). Unless stated otherwise, we use
\(\beta_{\mathrm{cb}}=0.96\).

\subsection{Details of Short-Term Angular-Band Filtering}
\label{subsec:detail_short_term_filtering}

The main text defines the over-aligned set
\(\mathcal H_{t,i}^{\mathrm{st}}\), the conflicting set
\(\mathcal C_{t,i}^{\mathrm{st}}\), and the resulting short-term correction.
Here we provide the weighting and scheduling details. The near-orthogonal safe
band is
\[
\mathcal S_{t,i}^{\mathrm{st}}
=
\{j:|\langle q_{t,i},\bar c_{t-1,j}\rangle|\le \epsilon_{\mathrm{st}}\}.
\]
Centroids in this set are left unchanged. For the risky sets, we use normalized
softmax weights
\[
\alpha_{t,ij}^{+}
=
\frac{
\exp\!\left(\tau_+\langle q_{t,i},\bar c_{t-1,j}\rangle\right)}
{\sum_{\ell\in\mathcal H_{t,i}^{\mathrm{st}}}
\exp\!\left(\tau_+\langle q_{t,i},\bar c_{t-1,\ell}\rangle\right)},
\qquad
j\in\mathcal H_{t,i}^{\mathrm{st}},
\]
\[
\alpha_{t,ij}^{-}
=
\frac{
\exp\!\left(\tau_-|\langle q_{t,i},\bar c_{t-1,j}\rangle|\right)}
{\sum_{\ell\in\mathcal C_{t,i}^{\mathrm{st}}}
\exp\!\left(\tau_-|\langle q_{t,i},\bar c_{t-1,\ell}\rangle|\right)},
\qquad
j\in\mathcal C_{t,i}^{\mathrm{st}}.
\]
If either risky set is empty, its corresponding projected component is set to
zero.

\paragraph{Adaptive conflict coefficient.}
We adapt \(\gamma_{\mathrm{st},t}\) according to the energy outside the safe
band. For each row,
\[
r_{t,i}
=
\frac{
\|z_{t,i}^{\mathrm{conf}}\|_2
+
\lambda_{\mathrm{hi}}\|z_{t,i}^{\mathrm{hi}}\|_2
}
{\|z_{t,i}\|_2+\varepsilon},
\qquad
r_t
=
\frac{1}{m}\sum_{i=1}^{m}r_{t,i}.
\]
We smooth this batch-level conflict signal by EMA,
\[
\bar r_t
=
\beta_r\bar r_{t-1}
+
(1-\beta_r)r_t,
\]
and set
\[
\gamma_{\mathrm{st},t}
=
\gamma_{\mathrm{st}}^{\min}
+
\left(\gamma_{\mathrm{st}}^{\max}
-
\gamma_{\mathrm{st}}^{\min}\right)
\frac{\bar r_t}{\bar r_t+\kappa_{\mathrm{st}}}.
\]
This increases suppression when more update energy lies in risky directions and
relaxes it when updates remain near the safe band. Unless stated otherwise, we
use fixed temperatures \(\tau_+,\tau_-\), EMA decay \(\beta_r=0.95\), and
sensitivity \(\kappa_{\mathrm{st}}=0.20\).

\subsection{Details of Long-Term Near-Orthogonal Protection}
\label{subsec:detail_long_term_protection}

The main text defines the active protection set
\(\mathcal B_{t,i}^{\mathrm{lt}}\), the protected component
\(z_{t,i}^{\mathrm{lt}}\), and the final protected update
\(\bar z_{t,i}\). Here we provide the remaining implementation details:
candidate selection, importance normalization, and centroid freezing.

\paragraph{Candidate set.}
At optimization step \(t\) of task \(\mathcal T_r\), the candidate set
\(\mathcal A_t^{\mathrm{lt}}\) is drawn from the frozen memory
\[
\mathcal M_{<r}^{\mathrm{lt}}
=
\bigcup_{\tau<r}
\mathcal M_{\tau}^{\mathrm{lt}},
\qquad
\mathcal M_{\tau}^{\mathrm{lt}}
=
\{(\bar c_{\mathcal T_\tau,k},\nu_{\mathcal T_\tau,k})\}_{k=1}^{K_\tau}.
\]
In practice, \(\mathcal A_t^{\mathrm{lt}}\) may be the full frozen bank or a
compact sampled subset for efficiency. When sampling is used, we sample
uniformly or according to normalized importance,
\[
p_{\mathcal T_\tau,k}
=
\frac{\nu_{\mathcal T_\tau,k}}
{\sum_{(\mathcal T_{\tau'},\ell)\in\mathcal M_{<r}^{\mathrm{lt}}}
\nu_{\mathcal T_{\tau'},\ell}+\varepsilon}.
\]

\paragraph{Importance normalization and the condition of Theorem~4}.
The weight \(\nu_{\mathcal T_\tau,k}\) controls the strength of suppression
along a frozen direction. To avoid scale imbalance across tasks, we normalize
the weights within each frozen task:
\[
\nu_{\mathcal T_\tau,k}
\leftarrow
\frac{\nu_{\mathcal T_\tau,k}}
{\sum_{\ell=1}^{K_\tau}\nu_{\mathcal T_\tau,\ell}+\varepsilon}.
\]

Each frozen centroid is a unit vector and the weights of one task sum to one,
so the operator of a single task,
$\Omega_\tau=\sum_k\nu_{\tau,k}\bar c_{\tau,k}\bar c_{\tau,k}^\top$, has trace
$1$ and hence $\lambda_{\max}(\Omega_\tau)\le1$. Summing over the
$T_{\mathrm{act}}$ frozen tasks that pass the activation test gives
$\lambda_{\max}(\Omega)\le T_{\mathrm{act}}$, with equality only if those
tasks froze the same direction with full weight. This bound alone would make
the condition of Theorem~4 depend on the number of tasks and on how aligned
their frozen directions are.

Rescaling removes this dependence. At each task boundary,
$\lambda_{\max}(\Omega)$ is computed exactly as the largest eigenvalue of the
$K\times K$ matrix $\mathrm{diag}(\sqrt\nu)\,\bar C\bar C^\top\mathrm{diag}(\sqrt\nu)$,
and the weights $\nu$ are scaled so that $\lambda_{\max}(\Omega)=r_{\mathrm{lt}}\le1$.
With $\gamma_{\mathrm{lt}}\le1$ the condition of Theorem~4 then holds for any
number of tasks. Subsampling the active set at a step only drops rank-one
terms, which cannot increase $\lambda_{\max}$, so the scale fixed at the
boundary remains valid at every step.

\paragraph{Centroid selection.}
At the end of task \(\mathcal T_\tau\), we select \(K_\tau\) centroids from the
online codebook using accumulated usage and diversity-adjusted importance. The
selected centroids are normalized, stored with their weights in
\(\mathcal M_\tau^{\mathrm{lt}}\), and kept fixed during later tasks.

\paragraph{Second-order long-term memory.}
In replay-free task sequences the frozen centroids are replaced by an
orthonormal basis $Q\in\mathbb R^{n\times r}$ built from the same update
stream, without random projection ($d=n$;
Section~\ref{subsubsec:two_capacities}). Within a task, each layer
accumulates the covariance of its short-term-filtered rows,
\[
\Psi_t=\rho\,\Psi_{t-1}+\sum_{i}\tilde z_{t,i}\tilde z_{t,i}^{\top},
\]
so that $\Psi$ describes the directions in use at the end of the task. At the
task boundary, the leading eigenvectors of the residual
$(I-QQ^\top)\Psi(I-QQ^\top)$ are appended to $Q$, and $\Psi$ is reset. In
later tasks every row is projected onto the complement of $\mathrm{span}(Q)$,
and the projection is applied once more after proximal lifting, so the update
reaching the weights is exactly orthogonal to $\mathrm{span}(Q)$ (Theorem~4
with $\Omega=QQ^\top$, $\gamma_{\mathrm{lt}}=1$). No activations or samples
are stored.

\section{Detailed Experimental Setup}
\label{sec:more_exps_settings_appendix}

We provide implementation details for all settings. All experiments are conducted on a cluster equipped with NVIDIA H100 and H200 GPUs and 256-core CPUs, using CUDA 13.0 and PyTorch 2.6. For computationally intensive tasks, we use multi-GPU training: 4$\times$H200 for language pretraining and 2$\times$H200 for continual fine-tuning. All remaining experiments run on a single GPU. Complete environment specifications and dependencies are provided in the attached code.

\subsection{Class-Imbalanced Learning}

\textbf{Datasets.} We construct a class-imbalanced variant of CIFAR-10 by designating five classes (4--8) as \emph{rare}, retaining only 10\% of their training samples while keeping all samples from the remaining classes. A ResNet-18 is trained from scratch for 50 epochs with batch size 128. We use a learning rate of $10^{-3}$ for both Adam and \method{}. For Muon, we find this learning rate too small for convergence and increase it to $10^{-2}$ to match the effective update scale of Adam and \method{}. Results are averaged over ten seeds.

\textbf{\method{} configuration.}
We maintain separate codebooks for backbone stages and the classifier head. For the backbone, we use codebook size 96, projection dimension 128, and slow/fast momentum coefficients $\mu_s = 0.97$, $\mu_f = 0.35$, with $K_s = K_f = 5$ Newton--Schulz iterations. The codebook EMA rate is $\beta_{\mathrm{ema}} = 0.95$, and spherical interpolation weight is $\xi = 0.25$. Short-term filtering uses $\epsilon_{\mathrm{st}} = 0.15$, $\gamma_{\mathrm{hi}} = 0.05$, and $\gamma_{\mathrm{st}} \in [0.10, 0.30]$ adapted via $\beta_r = 0.95$ and $\kappa_{\mathrm{st}} = 0.15$. The decorrelation coefficient is $\lambda_c = 0.10$ with $r = 8$ nearest neighbors. Inactive centroids are re-seeded every 50 optimizing steps at a reset threshold of 0.03. The proximal lifting coefficient is $\lambda_{\mathrm{prox}} = 1.0$ and the update scale is $0.1$ for both backbone and head. For the classifier head, we use a codebook size of 64 and a projection dimension of 128.

\subsection{Continual Visual Learning}

\subsubsection{Domain-Incremental Learning.}

\textbf{Datasets.} PACS~\citep{li2017deeper} contains 9,991 images of seven shared classes across four domains: Photo, Art Painting, Cartoon, and Sketch. All images are resized to $224 \times 224$ and split 80/20 into training and held-out test sets. We fine-tune an ImageNet-pretrained ResNet-18 with frozen batch-normalization statistics for 15 epochs per domain task, using a learning rate of $5 \times 10^{-5}$, batch size of 32, and gradient clipping at norm 1.0. Results are averaged over three randomly sampled domain orders and two seeds.
 
\textbf{\method{} configuration.}
We organize \method{}'s memory at the level of semantic network stages rather than individual tensors: separate codebooks are maintained for the stem and each residual stage, while a smaller independent codebook serves the classifier head. Centroids are shared within each stage, but random projections remain tensor-shape specific to accommodate varying dimensions. For the backbone, we use codebook size 64, projection dimension 128, and freeze up to $K_{\mathrm{frz}} = 21$ centroids per stage at each task boundary; for the classifier head, we use codebook size 32, projection dimension 64, and $K_{\mathrm{frz}} = 12$. The slow and fast momentum coefficients are $\mu_s = 0.95$ and $\mu_f = 0.35$ respectively, with $K_s = K_f = 5$ Newton--Schulz iterations and spherical interpolation weight $\xi = 0.3$. Short-term filtering uses $\epsilon_{\mathrm{st}} = 0.15$; long-term protection uses $\epsilon_{\mathrm{lt}} = 0.10$ and $\gamma_{\mathrm{lt}} = 0.6$ for the backbone ($0.4$ for the head). The codebook EMA rate is $\beta_{\mathrm{ema}} = 0.95$, with inactive centroids re-seeded every 50 steps at a reset threshold of $0.03$. Frozen centroids are selected via a drift-aware score combining usage, uniqueness, and movement from the pre-task snapshot, with a larger budget allocated to deeper stages where domain-specific drift is more pronounced. The proximal lifting coefficient is $\lambda_{\mathrm{prox}} = 1.0$, and the update scale is $0.1$ for both backbone and head.
 
\textbf{Continual learning baselines.}
When combining with continual learning methods, we use their default hyperparameters: EWC~\citep{kirkpatrick2017overcoming} with $\lambda = 5000$, online mode, and 1,024 Fisher samples; ER~\citep{chaudhry2019tiny} with buffer size 100, replay batch size 32, and $\alpha = 1.0$; DER++~\citep{buzzega2020dark} with buffer size 100, replay batch size 32, $\alpha = 0.1$, and $\beta = 0.5$.

\subsubsection{Class Incremental-Learning}

\paragraph{Class-Incremental Learning on CIFAR-10.}
We split CIFAR-10 into five sequential tasks with two classes each: $\{0,1\}$, $\{2,3\}$, $\{4,5\}$, $\{6,7\}$, $\{8,9\}$. All images are $32 \times 32$ with standard normalization. A ResNet-18 is trained from scratch for 20 epochs per task with learning rate $2 \times 10^{-3}$, batch size 128, and gradient clipping at norm 2.0. We use DER++~\citep{buzzega2020dark} as the base continual learning framework with a replay buffer of size 200, replay batch size 32, $\alpha = 0.1$, and $\beta = 0.5$, and replace only the optimizer among Adam, Muon, UPGD~\citep{elsayed2024addressing}, and \method{}. Results are averaged over five seeds. The component ladder of Table~\ref{tab:ladder} uses the same protocol.

\textbf{\method{} configuration.}
For class-incremental learning, \method{} organizes memory at the level of semantic network stages. In the backbone, codebooks are shared within each residual stage (stem, layer1--4), while random projections remain tensor-shape specific to handle varying dimensions within a stage. The classifier head uses two separate codebooks: one for the intermediate FC layer ($512 \times 512$) and a smaller one for the classifier output ($10 \times 512$, approximately one centroid per class). This three-tier design (backbone stages, intermediate head, classifier) allows each component to evolve at its natural granularity.

For the backbone, we use codebook size 96, projection dimension 256, slow and fast momentum $\mu_s = 0.97$, $\mu_f = 0.20$, with $K_s = 4$ and $K_f = 4$ Newton--Schulz iterations, and spherical interpolation weight $\xi = 0.20$. The codebook EMA rate is $\beta_{\mathrm{ema}} = 0.96$. Short-term filtering uses $\epsilon_{\mathrm{st}} = 0.25$, $\gamma_{\mathrm{hi}} = 0.01$, and $\gamma_{\mathrm{st}} \in [0.15, 0.40]$ adapted via $\beta_r = 0.95$ and $\kappa_{\mathrm{st}} = 0.15$. Long-term protection uses $\epsilon_{\mathrm{lt}} = 0.001$ and $\gamma_{\mathrm{lt}} = 1.0$ for both anti-aligned and aligned directions, with the importance weights rescaled so that $\lambda_{\max}(\Omega)\le1$ (Appendix~\ref{subsec:detail_long_term_protection}) and a maximum of 48 active frozen centroids sampled per step for efficiency. The decorrelation coefficient is $\lambda_c = 0.05$ with $r = 8$ nearest neighbors. Inactive centroids are re-seeded every 50 steps at a reset threshold of 0.10. The proximal lifting coefficient is $\lambda_{\mathrm{prox}} = 1.0$ and backbone update scale is $0.1$.

For the intermediate head FC ($512 \times 512$), we use codebook size 48, projection dimension 96, and freeze quota of 20 centroids per task. For the classifier ($10 \times 512$), we use codebook size 12, projection dimension 16, and freeze quota of 8 centroids per task, with $\gamma_{\mathrm{lt}} = 1.0$ and update scale $0.05$. The head \method{} learning rate is $5 \times 10^{-4}$, and the auxiliary AdamW learning rate for 1D parameters is also $5 \times 10^{-4}$.

At each task boundary, \method{} freezes a stage-dependent number of diverse centroids: 21 per backbone stage (stem through layer4), 20 for the intermediate head FC, and 8 for the classifier. Centroids are selected by an importance score that balances usage frequency, directional diversity, and drift from the task-start snapshot.

\paragraph{Task-Incremental Learning on CIFAR-100 (GPM protocol).}
We follow the official protocol of GPM~\citep{saha2021gradient}: CIFAR-100 is split into ten tasks of ten classes, the network is the five-layer AlexNet of GPM (three convolutions of 64, 128, and 256 channels with $4\times4$, $3\times3$, and $2\times2$ kernels, followed by two fully connected layers of width 2048) trained from scratch with a task-specific output head, task identity is given at test time, and no sample is replayed. Each task is trained for up to 200 epochs with batch size 64 and no data augmentation; the learning rate starts at $10^{-3}$ and is halved whenever the validation loss (5\% of the training data) has not improved for six epochs, down to $10^{-5}$, after which training stops. Following the official code, the batch-normalization affine parameters are frozen after the first task. GPM is run from its official code with its default energy threshold $0.97+0.003\tau$ and learning rate $10^{-2}$. We report the average accuracy over all ten tasks after the last task, over four seeds.

\textbf{\method{} configuration.}
This is the replay-free regime of Section~\ref{subsubsec:two_capacities}, so \method{} uses its second-order long-term memory, without random projection ($d=n$, i.e.\ $48$, $576$, $512$, $1024$, and $2048$ for the five layers; the head is task-private and keeps a codebook of size 12 with $d=16$). Each layer has its own codebook of size 12, 48, 48, 128, and 128. The optimizer uses $\mu_s=0.97$, $\mu_f=0.20$, $K_s=1$ and $K_f=2$ Newton--Schulz iterations, $\xi=0.1$, $\beta_{\mathrm{ema}}=0.995$, short-term filtering with $\epsilon_{\mathrm{st}}=0.25$, $\gamma_{\mathrm{hi}}=0.01$, $\gamma_{\mathrm{st}}\in[0.10,0.20]$, $\beta_r=0.95$, $\kappa_{\mathrm{st}}=0.15$, $\lambda_{\mathrm{prox}}=5$, re-seeding every 1000 steps at threshold 0.02, and update scale $0.02$ for the backbone ($0.01$ for the head). The second-order memory accumulates the covariance of the filtered rows with $\rho=0.9$ (a window of about ten updates). At each boundary at most $n/10$ directions per layer are added (5, 55, 49, 97, and 195), with a lifetime cap of $n/2$ per layer, or fewer if the basis already captures a fraction $0.97+0.003\tau$ of $\mathrm{tr}(\Psi)$; in our runs the per-task cap binds first, so the first five tasks fill the budget. Protection is the hard projection $\gamma_{\mathrm{lt}}=1$, applied in $\mathbb R^{n}$ and again after proximal lifting.

\subsection{Continual Fine-tuning}

\paragraph{Datasets.} We evaluate on Continual VisualQA, the VQA~v2~\citep{goyal2017making} benchmark of CL-MoE~\citep{huai2025cl} (200K images, 1.1M questions), split into ten reasoning-type tasks learned in sequence: recognition, location, judgment, commonsense, counting, action, color, type, subcategory, and causal reasoning. The base model is LLaVA-1.5-7B~\citep{liu2023visual} with a frozen CLIP ViT-L/14@336 encoder. We apply MoE-LoRA ($r{=}16$, $\alpha{=}16$, 4 experts) to the three MLP modules (\texttt{gate\_proj}, \texttt{up\_proj}, \texttt{down\_proj}). Each task trains for 1 epoch with learning rate $2 \times 10^{-4}$ (cosine, 3\% warmup), effective batch size 32, bfloat16, and sequence length 2048.

\textbf{\method{} configuration.}
With rank-16 LoRA, all gradient matrices have $\min(m,n){=}16$. We apply the full \method{} pipeline to every LoRA-A and LoRA-B parameter by projecting along its larger dimension, rather than falling back to AdamW for small tensors.

\method{} maintains 9 codebooks ($C{=}48$, $d{=}128$), one per combination of depth group (shallow/middle/deep, splitting the 32 LLaMA layers into thirds) and MLP module type. Gate, up, and down projections are separated because their gradient semantics differ (gating vs.\ expansion vs.\ compression); a shared codebook would freeze irrelevant directions. All 4 MoE experts at the same depth and module type contribute to a single codebook, enabling cross-expert knowledge transfer. Since LoRA-B is tall ($11008 {\times} 16$) while LoRA-A is wide ($16 {\times} 4096$), each codebook uses two shape-specific random projections (for $D{=}4096$ and $D{=}11008$)---18 projections in total.

The key hyperparameters: dual-stream momentum $\mu_s{=}0.96$, $\mu_f{=}0.16$ with 5/4 Newton--Schulz steps and SLERP blend $\xi{=}0.25$; short-term filter $\epsilon_{\mathrm{st}}{=}0.20$, $\gamma_{\mathrm{hi}}{=}{}0.01$, $\gamma_{\mathrm{st}} \in [0.15, 0.45]$; long-term protection $\epsilon_{\mathrm{lt}}{=}0.05$, $\gamma_{\mathrm{lt}}{=}1.0$; proximal coefficient $\lambda_{\mathrm{prox}}{=}0.8$; update scale $\sigma{=}0.2$.

At each task boundary, \method{} freezes 23 centroids per codebook, with a larger budget allocated to deeper layers, selected by a score balancing usage, directional uniqueness, and drift from the task-start snapshot.

\subsection{Language Pretraining}

\paragraph{Datasets.}
We pretrain GPT-2 Small (125M parameters) from scratch on OpenWebText~\citep{Gokaslan2019OpenWeb} (${\sim}$9B training tokens, 4.4M validation tokens) using the nanoGPT implementation~\citep{Karpathy2022}. We remove linear-layer biases, use GeLU activations, set dropout to $0.0$, replace learned positional embeddings with RoPE~\citep{su2024roformer}, and adopt a warmup-stable learning rate schedule~\citep{hu2024minicpm}. All models are trained for 100K steps on approximately 49.2B tokens, with context length 1024, 2K warmup steps, and an effective batch size of 60 sequences. Architecture, data order, token budget, and all non-optimizer hyperparameters are kept identical across methods.

\textbf{Baselines.}
We compare \method{} with Adam~\citep{loshchilov2018decoupled} and Muon~\citep{jordan2024muon}. For Muon-style optimizers, we follow the standard hybrid setup: the orthogonalization-based optimizer is applied to matrix-valued parameters (2D weights), while all remaining parameters (embeddings, biases, layer norms) are updated with Adam. We adopt the same hybrid setup for \method{}. For Adam, we use $(\beta_1, \beta_2) = (0.9, 0.95)$; for Muon, $\beta = 0.95$. Weight decay is set to $0.1$ and learning rate to $6 \times 10^{-4}$ for all methods.

\textbf{\method{} configuration.}
For GPT-2, \method{} organizes codebooks by parameter shape, yielding four codebook types shared across all transformer blocks: \texttt{c\_attn} ($768 \times 2304$), \texttt{c\_proj} ($768 \times 768$), \texttt{mlp\_fc} ($768 \times 3072$), and \texttt{mlp\_proj} ($3072 \times 768$). Parameters with the same shape share a single codebook, and attention vs.\ MLP parameters are distinguished via module-type tags. We use codebook size 256 and projection dimension 512, with slow and fast momentum coefficients $\mu_s = 0.95$ and $\mu_f = 0.15$, $K_s = K_f = 5$ Newton--Schulz iterations, and spherical interpolation weight $\xi = 0.25$. The codebook EMA rate is $\beta_{\mathrm{ema}} = 0.96$. Short-term filtering uses $\epsilon_{\mathrm{st}} = 0.20$, $\gamma_{\mathrm{hi}} = 0.001$, and $\gamma_{\mathrm{st}} \in [0.15, 0.30]$ adapted via $\beta_r = 0.95$ and $\kappa_{\mathrm{st}} = 0.15$. The decorrelation coefficient is $\lambda_c = 0.05$ with $r = 8$ nearest neighbors, and inactive centroids are re-seeded every 200 steps at a reset threshold of 0.10. The proximal lifting coefficient is $\lambda_{\mathrm{prox}} = 1.0$, and the final update scaling follows $\gamma_t = \sigma \sqrt{mn} / (\|\hat{O}_t\|_F + \varepsilon)$ with $\sigma = 0.2$. As this is a single-task pretraining setting, long-term protection is inactive. Training is distributed across 4$\times$H200 GPUs with consistent random projection seeds across ranks.

\section{More Experimental Results}
\label{sec:more_exps_appendix}

\subsection{Class-Imbalanced Learning on CIFAR-100}
We repeat the class-imbalanced protocol of Section~\ref{subsection:class-imbalanced-learning} on CIFAR-100: half of the classes are designated rare and keep 10\% of their training samples (50 instead of 500 images per class), ResNet-18 is trained from scratch for 50 epochs, and the optimizer settings are those of the CIFAR-10 experiment without further tuning, five seeds. Table~\ref{tab:imbalanced_cifar100} shows the same pattern as at ten classes: Muon loses rare-class accuracy sharply, and \method{} restores it ($+6.2$ over Muon) while also surpassing Adam on all three metrics.

\begin{table}[t]
\centering
\caption{\textbf{Class-imbalanced CIFAR-100.} Half of the classes rare at 10\% of their samples; ResNet-18 from scratch, 50 epochs, five seeds. Forgetting is the drop from the best-so-far rare-class accuracy.}
\label{tab:imbalanced_cifar100}
\small
\setlength{\tabcolsep}{8pt}
\renewcommand{\arraystretch}{1.12}
\begin{tabular}{@{}lccc@{}}
\toprule
\rowcolor{sectionblue}
\textbf{Optimizer} & \textbf{Overall acc. (\%)} $\uparrow$ & \textbf{Rare acc. (\%)} $\uparrow$ & \textbf{Rare forgetting (\%)} $\downarrow$ \\
\midrule
\rowcolor{standard} Muon & $34.94 \pm 0.44$ & $13.81 \pm 0.48$ & $2.57 \pm 0.85$ \\
\rowcolor{standard} Adam & $39.18 \pm 0.43$ & $19.44 \pm 1.34$ & $2.43 \pm 0.77$ \\
\rowcolor{copehl} \textbf{\textcolor{copegreen}{\method{}}} & \textcolor{copegreen}{$\mathbf{40.60 \pm 0.36}$} & \textcolor{copegreen}{$\mathbf{20.03 \pm 0.21}$} & \textcolor{copegreen}{$\mathbf{2.32 \pm 0.69}$} \\
\bottomrule
\end{tabular}
\end{table}

\subsection{Continual Visual Learning}
\subsubsection{Combining Optimizers with Other Continual Learning Methods}

We further examine whether \method{} can complement existing continual learning
techniques. We evaluate PACS Domain-IL with two representative approaches:
EWC~\citep{kirkpatrick2017overcoming}, a regularization-based method, and DER++, a replay-and-distillation method.
For each method, we replace the optimizer while keeping the continual learning
setup fixed.

\begin{table}[t]
\centering
\caption{\textbf{Combining optimizers with continual learning methods on PACS Domain-IL.}
Results are averaged over three randomly selected domain orders and two seeds. AP denotes Average Performance, and AF denotes Average Forgetting. Best results are in bold and second-best results are underlined.}
\label{tab:pacs_dil_results}
\small
\setlength{\tabcolsep}{3.8pt}
\renewcommand{\arraystretch}{1.12}
\resizebox{\linewidth}{!}{%
\begin{tabular}{@{}llccccc@{}}
\toprule
\rowcolor{sectionblue}
\textbf{CL Method} & \textbf{Optimizer} & \textbf{Memory} & \textbf{Mem. Usage (MB)} $\downarrow$ & \textbf{Train Time (m)} $\downarrow$ & \textbf{AP (\%)} $\uparrow$ & \textbf{AF (\%)} $\downarrow$ \\
\midrule

\rowcolor{standard}
EWC & Adam & Fisher & 85.3 & \textbf{8.44} & $83.94 \pm 4.26$  & $13.90 \pm 6.59$  \\
\rowcolor{standard}
EWC & Muon  & Fisher & 85.3 & \underline{9.11} & \underline{$88.57 \pm 2.05$} & \underline{$7.93 \pm 2.65$} \\
\rowcolor{standard}
EWC & CBP   & Fisher & 85.3 & 12.85 & $84.42 \pm 1.06$ & $13.56 \pm 2.85$ \\
\rowcolor{standard}
EWC & FIRE  & Fisher & 85.3 & 11.46 & $80.44 \pm 4.60$ & $14.76 \pm 3.68$ \\
\rowcolor{copehl}
EWC & \textbf{\textcolor{copegreen}{\method{}}} & Fisher + Centroids 
& \textcolor{copegreen}{$\underline{85.9}$} & \textcolor{copegreen}{10.98} & \textcolor{copegreen}{$\mathbf{88.81 \pm 3.02}$}  & \textcolor{copegreen}{$\mathbf{6.26 \pm 2.01}$}  \\

\cmidrule(lr){1-7}

\rowcolor{standard}
DER++ & Adam & Buffer + logits & 57.4 & \textbf{8.37} & $92.21 \pm 1.18$ & $4.51 \pm 0.64$ \\
\rowcolor{standard}
DER++ & Muon  & Buffer + logits & 57.4 & \underline{8.98} & \underline{$92.86 \pm 1.48$} & \underline{$3.08 \pm 1.01$} \\
\rowcolor{standard}
DER++ & CBP   & Buffer + logits & 57.4 & 13.90 & $91.87 \pm 1.42$ & $4.66 \pm 0.55$ \\
\rowcolor{standard}
DER++ & FIRE  & Buffer + logits & 57.4 & 11.63 & $88.12 \pm 2.94$ & $6.85 \pm 0.25$  \\
\rowcolor{copehl}
DER++ & \textbf{\textcolor{copegreen}{\method{}}} & Buffer + logits + Centroids 
& \textcolor{copegreen}{$\underline{58.0}$} & \textcolor{copegreen}{11.02} & \textcolor{copegreen}{$\mathbf{92.91 \pm 2.07}$} & \textcolor{copegreen}{$\mathbf{2.60 \pm 1.16}$} \\

\bottomrule
\end{tabular}%
}
\end{table}

Table~\ref{tab:pacs_dil_results} shows that \method{} consistently improves both
regularization-based and replay-based continual learning. With EWC, \method{}
achieves the highest AP and reduces AF from \(7.93\%\) with Muon to \(6.26\%\).
With DER++, \method{} attains the best AP and further lowers AF from \(3.08\%\)
to \(2.60\%\). These gains indicate that directional memory is complementary to
existing continual learning mechanisms: EWC and DER++ impose task-level
constraints, while \method{} preserves useful update directions at the optimizer
level.

\subsubsection{Additional Task-Order Results}

To further assess the robustness of \method{}, we report additional Class-IL
results on CIFAR-10 under different task orders. Beyond the main order used in
Fig.~\ref{fig:class_il_main}, we evaluate two alternative sequences:
\([[0,1],[8,9],[3,5],[4,7],[2,6]]\) and
\([[3,7,9,0],[1,5],[4,8],[2],[6]]\), covering both balanced and imbalanced
class splits. Table~\ref{tab:task_order_sensitivity} shows that \method{}
consistently outperforms Adam, Muon, and UPGD across task orders while achieving
lower forgetting. This suggests that the gains are not tied to a favorable class
sequence, but reflect the ability of directional memory to preserve
task-relevant updates across different incremental learning trajectories.

\begin{table}[t]
\centering
\caption{\textbf{Additional CIFAR-10 Class-IL results across different task orders.}
AP denotes Average Performance, and AF denotes Average Forgetting. Results are averaged over five random seeds. Best results are in bold and second-best results are underlined.}
\label{tab:task_order_sensitivity}
\scriptsize
\setlength{\tabcolsep}{4.2pt}
\renewcommand{\arraystretch}{1.10}
\resizebox{0.68\linewidth}{!}{%
\begin{tabular}{@{}llcc@{}}
\toprule
\rowcolor{sectionblue}
\textbf{Task Order} & \textbf{Optimizer} & \textbf{AP (\%)} $\uparrow$ & \textbf{AF (\%)} $\downarrow$ \\
\midrule

\rowcolor{sectionblue!45}
\multicolumn{4}{@{}l}{\emph{Order 1: balanced two-class increments}} \\
\rowcolor{standard}
\multirow{4}{*}{$[[0,1],[8,9],[3,5],[4,7],[2,6]]$}
& Adam & \underline{$38.54 \pm 8.41$} & \underline{$52.21 \pm 6.40$} \\
\rowcolor{standard}
& Muon & $26.62 \pm 4.21$ & $65.85 \pm 3.14$ \\
\rowcolor{standard}
& UPGD & $16.37 \pm 0.58$ & $63.96 \pm 0.93$ \\

\cmidrule(lr){1-4}
\rowcolor{copehl}
& \textbf{\textcolor{copegreen}{\method{}}} 
& \textcolor{copegreen}{$\mathbf{43.83 \pm 5.75}$} 
& \textcolor{copegreen}{$\mathbf{50.32 \pm 4.11}$}  \\

\midrule

\rowcolor{sectionblue!45}
\multicolumn{4}{@{}l}{\emph{Order 2: imbalanced class increments}} \\
\rowcolor{standard}
\multirow{4}{*}{$[[3,7,9,0],[1,5],[4,8],[2],[6]]$}
& Adam & $41.16 \pm 7.01$ & \underline{$52.94 \pm 6.47$} \\
\rowcolor{standard}
& Muon & $28.46 \pm 2.54$ & $66.54 \pm 3.49$ \\
\rowcolor{standard}
& UPGD & $20.04 \pm 0.08$ & $72.33 \pm 0.34$ \\
\cmidrule(lr){1-4}
\rowcolor{copehl}
& \textbf{\textcolor{copegreen}{\method{}}} 
& \textcolor{copegreen}{$\mathbf{44.45 \pm 5.01}$} 
& \textcolor{copegreen}{$\mathbf{52.71 \pm 4.97}$}  \\

\bottomrule
\end{tabular}%
}
\end{table}

\subsubsection{Class-Incremental Learning on CIFAR-100}
\label{subsec:class_il_cifar100}
We extend the Class-IL protocol to CIFAR-100 with ten sequential tasks of ten classes, twice the length of the CIFAR-10 setting: ResNet-18 trained from scratch for 50 epochs per task, the same class sequence and DER++ replay buffer of 500 samples for every method, five seeds, and no data augmentation, which lowers absolute accuracy uniformly across rows. Besides Adam and Muon we add a comparable-budget control, RP-OGD: it runs on \method{}'s substrate (same random projection, dimension, number of stored directions, and proximal lifting) but uses the stored directions by hard orthogonal projection, as OGD does, instead of codebook consolidation and calibrated correction. Table~\ref{tab:class_il_cifar100} reports the directional-memory footprint; the shared replay buffer and optimizer moments are excluded. A comparison with GPM under its own protocol is given in Appendix~\ref{subsec:task_il_results}.

\method{} improves on Muon by $+4.4$ AP and $-4.2$ AF at $1.86$\,MB and reaches parity with replay-only Adam ($20.9$ versus $21.0$ AP). RP-OGD is the direct test of what the memory contributes: with the same stored directions, hard projection gives no retention benefit over memory-less Muon ($70.2$ versus $70.2$ AF), whereas \method{}'s codebook consolidation and calibrated correction give $+4.7$ AP and $-4.1$ AF (paired). The gain comes from how the directional memory is built and applied, not from its existence.

\begin{table}[t]
\centering
\caption{\textbf{Class-incremental learning on CIFAR-100.} Ten tasks, ResNet-18 from scratch, DER++ base with a 500-sample buffer shared by all rows, five seeds. Memory counts the directional memory only. RP-OGD: comparable-budget control on \method{}'s substrate with hard projection in place of codebook consolidation and calibrated correction.}
\label{tab:class_il_cifar100}
\small
\setlength{\tabcolsep}{5pt}
\renewcommand{\arraystretch}{1.12}
\begin{tabular}{@{}llccc@{}}
\toprule
\rowcolor{sectionblue}
\textbf{Optimizer} & \textbf{Directional memory} & \textbf{Memory} $\downarrow$ & \textbf{AP (\%)} $\uparrow$ & \textbf{AF (\%)} $\downarrow$ \\
\midrule
\rowcolor{standard} Adam & none & -- & $21.03 \pm 1.96$ & $66.22 \pm 2.83$ \\
\rowcolor{standard} Muon & none & -- & $16.58 \pm 0.70$ & $70.24 \pm 0.99$ \\
\rowcolor{standard} RP-OGD & projected directions & 1.02\,MB & $16.26 \pm 1.60$ & $70.16 \pm 1.18$ \\
\rowcolor{copehl} \textbf{\textcolor{copegreen}{\method{}}} & projected codebook & \textcolor{copegreen}{1.86\,MB} & \textcolor{copegreen}{$\mathbf{20.94 \pm 1.25}$} & \textcolor{copegreen}{$\mathbf{66.06 \pm 1.47}$} \\
\bottomrule
\end{tabular}
\end{table}
\subsubsection{Task-Incremental Learning on CIFAR-100}
\label{subsec:task_il_results}

Table~\ref{tab:task_il_cifar100} compares \method{} with second-order memory against GPM under GPM's own protocol (Appendix~\ref{sec:more_exps_settings_appendix}). \method{} is better on every one of the four paired seeds, at comparable memory and without storing any sample: the long-term basis takes $11.7$\,MB and the online codebook $1.8$\,MB, against $12.9$\,MB for GPM's bases. Raising the lifetime cap from $n/2$ to $0.6n$ or $0.95n$ ($14.0$ and $22.1$\,MB) increases the accuracy by less than one point.

\begin{table}[t]
\centering
\caption{\textbf{Task-incremental learning on CIFAR-100 under the GPM protocol.} AlexNet, ten tasks, no replay, task identity at test time. Average accuracy over the ten tasks after the last task, mean $\pm$ std over four seeds. Memory counts the across-task store only (GPM: its bases; \method{}: the basis $Q$), with the online codebook listed separately.}
\label{tab:task_il_cifar100}
\small
\setlength{\tabcolsep}{8pt}
\renewcommand{\arraystretch}{1.12}
\begin{tabular}{@{}lcc@{}}
\toprule
\rowcolor{sectionblue}
\textbf{Method} & \textbf{Memory (MB)} $\downarrow$ & \textbf{AP (\%)} $\uparrow$ \\
\midrule
\rowcolor{standard}
GPM~\citep{saha2021gradient} & 12.9 & $70.67 \pm 0.46$ \\
\rowcolor{copehl}
\textbf{\textcolor{copegreen}{\method{}}} (second-order memory, $d=n$) & \textcolor{copegreen}{11.7 + 1.8} & \textcolor{copegreen}{$\mathbf{72.05 \pm 1.33}$} \\
\bottomrule
\end{tabular}
\end{table}

\subsection{Zero-shot OOD Generalization}

We further evaluate whether \method{} improves generalization to unseen task
types. We finetune LLaVA-1.5-7B with a LoRA-MoE architecture on the first seven
VQA tasks, then evaluate the resulting checkpoint zero-shot on the remaining
three held-out tasks. As shown in Table~\ref{tab:appendix_zero_shot_ood}, \method{}
consistently outperforms AdamW and Muon, suggesting that directional memory
helps preserve transferable directions for OOD generalization.

\begin{table}[t]
\centering
\caption{\textbf{Zero-shot OOD generalization.}
LLaVA-1.5-7B with LoRA-MoE is trained on the first seven VQA task types and evaluated on the last three held-out types without further finetuning. Best results are in bold and second-best results are underlined.}
\label{tab:appendix_zero_shot_ood}
\small
\setlength{\tabcolsep}{9pt}
\renewcommand{\arraystretch}{1.12}
\begin{tabular}{lccc}
\toprule
\textbf{Optimizer} & \textbf{Typ.} & \textbf{Sub.} & \textbf{Cau.} \\
\midrule
AdamW & {57.06} & 57.98 & 26.73 \\
Muon  & \underline{57.80} & \underline{58.35} & \underline{29.03} \\
\arrayrulecolor{softgold}\midrule[1.1pt]\arrayrulecolor{black}
\rowcolor{softgoldrow}
\textbf{\method{}} & \textbf{59.95} & \textbf{60.47} & \textbf{29.88} \\
\bottomrule
\end{tabular}
\end{table}

\subsection{Continual Instruction Tuning on Super-NaturalInstructions}
\label{subsec:superni}
We fine-tune a frozen LLaMA-2-7B with LoRA-MoE (8 experts, rank 4, on all attention and MLP projections) on eight Super-NaturalInstructions tasks in sequence (generation, extraction, classification, and dialogue; task ids in the order of Table~\ref{tab:superni}), five epochs per task. After the last task every task is evaluated again with ROUGE-L. The three LoRA-MoE rows share the architecture and differ only in the optimizer. InfLoRA~\citep{liang2024inflora} and O-LoRA~\citep{wang2023orthogonal} use the same backbone, task order, and training budget with their own mechanism: both freeze the adapter branches of previous tasks and add a new, near-orthogonal branch for each task.

\method{} reaches the highest AP of all methods. Within the same architecture it improves on Muon by $+1.8$ AP and $-2.2$ AF, and on AdamW by $+3.4$ AP and $-1.4$ AF. InfLoRA and O-LoRA have lower AF, but for an architectural reason: with frozen branches, old tasks cannot be overwritten, so their forgetting is bounded before the optimizer plays any role, and this protection costs plasticity (InfLoRA has the lowest AP, below plain AdamW). The first task (1572) shows both effects: the two branch-freezing methods keep $28$ to $36$ ROUGE-L on it, the optimizer rows $8$ to $16$, with \method{} the best of the three. \method{} therefore trades a small amount of forgetting for accuracy rather than minimizing forgetting, and the two approaches are complementary: as an optimizer, \method{} can be used on top of InfLoRA or O-LoRA.

\begin{table}[t]
\centering
\caption{\textbf{Continual instruction tuning on Super-NaturalInstructions.} ROUGE-L on each of the eight tasks after the last task, in learning order (task ids), with AP and AF. Frozen LLaMA-2-7B; optimizer rows share the LoRA-MoE architecture.}
\label{tab:superni}
\small
\setlength{\tabcolsep}{3.4pt}
\renewcommand{\arraystretch}{1.12}
\resizebox{\linewidth}{!}{%
\begin{tabular}{@{}lcccccccccc@{}}
\toprule
\rowcolor{sectionblue}
\textbf{Method} & \textbf{1572} & \textbf{363} & \textbf{1290} & \textbf{181} & \textbf{002} & \textbf{1510} & \textbf{639} & \textbf{1729} & \textbf{AP} $\uparrow$ & \textbf{AF} $\downarrow$ \\
\midrule
\rowcolor{standard} LoRA-MoE + AdamW & 8.55 & 89.00 & 25.67 & 65.33 & 61.81 & 98.33 & 12.21 & 15.50 & 47.05 & 5.24 \\
\rowcolor{standard} LoRA-MoE + Muon & 13.19 & 83.00 & 25.52 & 58.05 & 81.22 & 99.67 & 10.63 & 17.55 & 48.60 & 6.12 \\
\rowcolor{standard} InfLoRA & 28.31 & 87.00 & 16.32 & 60.51 & 65.70 & 98.69 & 7.03 & 14.96 & 47.32 & 1.81 \\
\rowcolor{standard} O-LoRA & 36.22 & 92.00 & 25.80 & 49.51 & 68.97 & 99.13 & 11.29 & 17.87 & 50.10 & 1.23 \\
\rowcolor{copehl} \textbf{\textcolor{copegreen}{LoRA-MoE + \method{}}} & 16.11 & 92.00 & 27.58 & 61.30 & 76.32 & 98.67 & 13.92 & 17.47 & \textcolor{copegreen}{$\mathbf{50.42}$} & 3.89 \\
\bottomrule
\end{tabular}%
}
\end{table}

\section{More Analyses and Ablation Studies}
\label{sec:more_analyses_and_ablation_studies}
\subsection{Components Analysis}
\label{appendix:components}

\textbf{Variants.}
We compare \method{} with versions of itself in which some mechanisms of Section~\ref{sec:Method} are switched off. All variants keep the full method's update normalization, layer partitioning, learning rates, and capacity $(d, C)$, so a difference between two rows is due to the mechanism that separates them.
\begin{itemize}
\item \emph{\method{}-Base}: orthogonalized momentum only. One slow stream ($\xi=0$), no short-term filtering ($\gamma_{\mathrm{st}}=\gamma_{\mathrm{hi}}=0$), no long-term protection ($\gamma_{\mathrm{lt}}=0$), no codebook decorrelation ($\lambda_c=0$). It differs from Muon only in these conventions.
\item \emph{\method{}-Min}: Base plus one projected codebook and conflict removal with a constant coefficient ($\gamma_{\mathrm{st}}=0.4$). This is the smallest version of the method: orthogonalization and one directional memory.
\item \emph{\method{}-Min+LT}: Min plus long-term protection with the hard setting $\gamma_{\mathrm{lt}}=1$.
\item \emph{\method{}} (full): Min+LT plus the fast stream with SLERP fusion, the adaptive short-term coefficient, the over-alignment deflation $\gamma_{\mathrm{hi}}$, and codebook decorrelation.
\end{itemize}
Table~\ref{tab:ladder} has two parts: the rows above the rule start from Muon and add these mechanisms one at a time; the rows below the rule start from the full method and remove one mechanism at a time.

\textbf{Results.}
Table~\ref{tab:ladder} reports Class-IL CIFAR-10 (DER++ base, protocol of Appendix~\ref{sec:more_exps_settings_appendix}, five paired seeds). Reading it top-down: \method{}-Base is already $9.9$ AP above Muon, which is the effect of the conventions alone, before any memory is used. The projected codebook adds $2.1$ AP and removes $1.9$ AF; long-term protection adds another $2.0$ AP and removes $2.2$ AF; the remaining mechanisms of the full method add $1.5$ AP and remove $4.6$ AF. Reading it bottom-up gives the same ordering: removing long-term protection from the full method costs $5.3$ AP and $8.6$ AF, removing short-term filtering costs $3.5$ AP and $6.7$ AF, and removing orthogonalization costs $2.0$ AP and $4.8$ AF. In this setting the directional memory, not the orthogonalization, carries most of the gain; the role of orthogonalization is the spectral equalization of Proposition~6, which matters most in the imbalanced and pretraining settings.

\method{}-Min+LT is the configuration we recommend when simplicity matters. Its settings are the capacity $(d, C, K_{\mathrm{frz}})$, the momentum and EMA constants, and two fixed strengths, $\gamma_{\mathrm{st}}=0.4$ and $\gamma_{\mathrm{lt}}=1$; it keeps most of the full method's performance ($44.88$ against $46.35$ AP), and the remaining mechanisms can be added when a setting calls for them.

\begin{table}[t]
\centering
\caption{\textbf{Component ladder on Class-IL CIFAR-10.} DER++ base, five paired seeds. Rows above the rule add one mechanism at a time, starting from Muon; rows below remove one mechanism from the full method. AP: Average Performance; AF: Average Forgetting.}
\label{tab:ladder}
\small
\setlength{\tabcolsep}{6pt}
\renewcommand{\arraystretch}{1.12}
\begin{tabular}{@{}lcc@{}}
\toprule
\rowcolor{sectionblue}
\textbf{Variant} & \textbf{AP (\%)} $\uparrow$ & \textbf{AF (\%)} $\downarrow$ \\
\midrule
\rowcolor{standard}
Muon & $30.95 \pm 4.21$ & $61.61 \pm 4.56$ \\
\rowcolor{standard}
\method{}-Base (orthogonalized momentum only) & $40.82 \pm 3.02$ & $57.02 \pm 3.19$ \\
\rowcolor{standard}
\method{}-Base + slow/fast streams & $41.44 \pm 3.20$ & $56.24 \pm 3.53$ \\
\rowcolor{standard}
\method{}-Min (Base + projected codebook, conflict removal) & $42.90 \pm 2.76$ & $55.12 \pm 2.89$ \\
\rowcolor{standard}
\method{}-Min+LT (+ long-term protection) & $44.88 \pm 2.88$ & $52.88 \pm 2.77$ \\
\midrule
\rowcolor{standard}
\method{} without long-term protection & $41.02 \pm 4.62$ & $56.86 \pm 4.68$ \\
\rowcolor{standard}
\method{} without short-term filtering & $42.85 \pm 1.16$ & $54.94 \pm 1.69$ \\
\rowcolor{standard}
\method{} without orthogonalization & $44.40 \pm 3.56$ & $53.04 \pm 3.33$ \\
\rowcolor{copehl}
\textbf{\textcolor{copegreen}{\method{}}} (full) & \textcolor{copegreen}{$\mathbf{46.35 \pm 3.54}$} & \textcolor{copegreen}{$\mathbf{48.27 \pm 3.88}$} \\
\bottomrule
\end{tabular}
\end{table}

\textbf{Capacity.}
Table~\ref{tab:capacity} halves or doubles the projected dimension $d$ or the codebook size $C$, one at a time, around the default. AP moves by at most $1.4$ points, less than one standard deviation, so the method is not sensitive to its capacity.

\begin{table}[t]
\centering
\caption{\textbf{Capacity sensitivity on Class-IL CIFAR-10.} Projected dimension $d$ and codebook size $C$ varied one at a time around the default ($d=256$, $C=96$); five seeds.}
\label{tab:capacity}
\small
\setlength{\tabcolsep}{8pt}
\renewcommand{\arraystretch}{1.12}
\begin{tabular}{@{}lcc@{}}
\toprule
\rowcolor{sectionblue}
\textbf{Capacity} & \textbf{AP (\%)} $\uparrow$ & \textbf{AF (\%)} $\downarrow$ \\
\midrule
\rowcolor{standard} $d/2$ & $44.94 \pm 2.83$ & $52.24 \pm 3.48$ \\
\rowcolor{standard} $2d$ & $45.21 \pm 3.78$ & $51.18 \pm 3.89$ \\
\rowcolor{standard} $C/2$ & $45.01 \pm 2.92$ & $51.33 \pm 3.31$ \\
\rowcolor{standard} $2C$ & $45.14 \pm 3.61$ & $51.16 \pm 3.87$ \\
\rowcolor{copehl} Default $(d, C)$ & $46.35 \pm 3.54$ & $48.27 \pm 3.88$ \\
\bottomrule
\end{tabular}
\end{table}

\subsection{Short-Term Forgetting Analysis}
\label{appendix:short-term-forgetting}

Fig.~\ref{fig:short-term-forgetting} instruments the class-imbalanced training loop (ResNet-18, CIFAR-10, 5 rare classes at 10\% data) with three per-step diagnostics, computed every 10 steps.

\paragraph{Panel~(a): Gradient conflict.}
At each logged step, we compute separate loss gradients for the dominant and rare samples within the current mini-batch and measure their cosine similarity. The cosine remains negative (${\approx}{-}0.1$) throughout training, indicating that the two groups consistently pull the model in opposing directions. This is the precondition for short-term forgetting: any update that benefits one group risks harming the other.

\paragraph{Panel~(b): Directional bias.}
We measure how much the combined gradient (the actual update direction) aligns with each group by projecting it onto the dominant and rare gradient directions. During steps 0--200, the projection onto the dominant group is high (${\sim}0.85$) while the rare group is low (${\sim}0.3$): the optimizer primarily serves the dominant group. After step ${\sim}$200, the pattern reverses---dominant samples have largely converged, their per-sample gradients shrink, and the rare group's larger gradients take over the combined update direction. This mirrors the subspace rotation described in Corollary 2: whichever group does not control the dominant subspace of $B_t$ has its directions suppressed.

\paragraph{Panel~(c): Direct forgetting evidence.}
To confirm that gradient conflict translates into actual loss degradation, we measure $\Delta L = L(\theta_{t+1}) - L(\theta_t)$ on a \emph{fixed held-out probe} of 256 samples per group, evaluated before and after each parameter update. During the dominant-serving phase (steps 0--200), the rare group's validation loss increases by up to $+0.10$ per step---direct evidence that the suppressed group is being harmed. After the directional bias reverses, the dominant group exhibits a persistent positive $\Delta L$, confirming that short-term forgetting shifts between groups as the optimization trajectory evolves. The use of held-out data rules out batch-level memorization as an explanation.

\subsection{Training Behaviors}
\label{subsec:training_curve_appendix}

\begin{figure}[t]
    \centering
    \includegraphics[width=\textwidth]{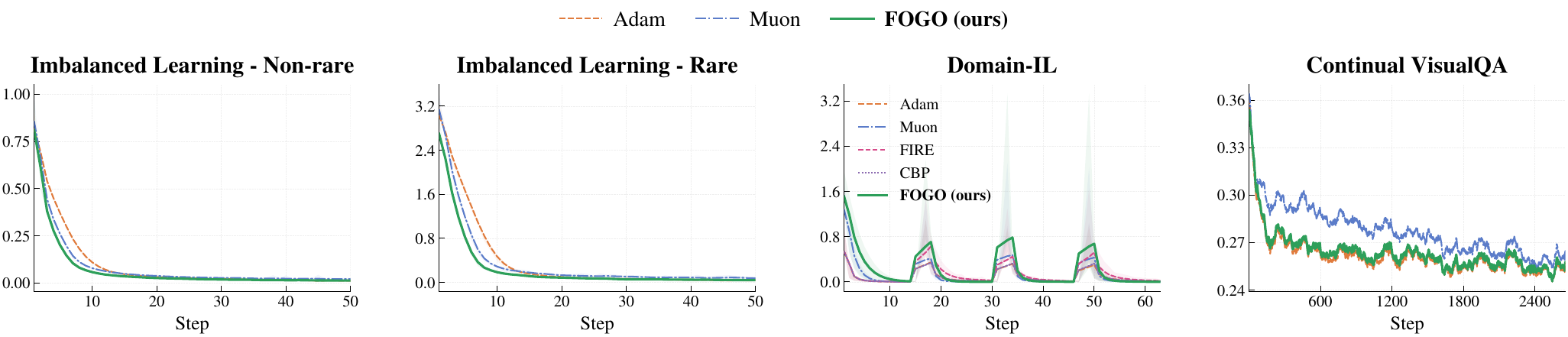}
    \caption{Training loss curves across four continual learning settings. From left to right: class-imbalanced learning on non-rare (dominant) and rare (non-dominant) classes, Domain-IL, and continual VisualQA with LoRA-MoE on LLaVA-1.5-7B. \method{} consistently converges faster and reaches a lower final loss than Adam and Muon across all settings. In Domain-IL, \method{} also outperforms specialized continual learning methods, including FIRE and CBP, and recovers rapidly after each domain shift.}

    \label{fig:training_curves_appendix}
\end{figure}

To comprehensively study the training dynamics of \method{}, we include the training loss curves across three settings in Fig.~\ref{fig:training_curves_appendix}.

\paragraph{Class-Imbalanced Learning.}
The first two panels of Fig.~\ref{fig:training_curves_appendix} show the per-class training loss on the non-rare and rare partitions, respectively. \method{} converges noticeably faster than both Adam and Muon within the first 5 steps and maintains a consistently lower loss throughout training. The advantage is especially pronounced on the rare classes, where the loss gap between \method{} and the baselines remains significant even after convergence, indicating that the gradient correction mechanism of \method{} is particularly effective for under-represented classes.

\paragraph{Domain-Incremental Learning.}
The third panel presents the training loss under domain-incremental learning, where the model encounters a sequence of domains. All methods exhibit periodic loss spikes at domain boundaries, as expected. However, \method{} recovers substantially faster after each transition compared to both standard optimizers (Adam, Muon) and dedicated continual learning baselines (FIRE, CBP), demonstrating its robustness to distributional shifts.

\paragraph{Continual VisualQA.}
The rightmost panel shows the training loss when integrating \method{} into LoRAMoE with LLaVA-1.5-7B for continual visual question answering. Despite the increased model complexity, \method{} achieves a lower and more stable training loss throughout, confirming that its benefits generalize beyond standard classification to multi-modal generative settings.


\subsection{Gradient Structure Preservation under Random Projection}
\label{subsec:projection-fidelity}

A central design choice in \method{} is the use of a random Rademacher projection $\Pi \in \mathbb{R}^{n \times d}$ to map each layer's gradient matrix from the original parameter space ($\mathbb{R}^n$) to a lower-dimensional codebook space ($\mathbb{R}^d$, $d \ll n$). For the downstream codebook operations to be meaningful, this projection must preserve the pairwise geometric structure of gradient directions---if two gradient rows are aligned or conflicting in the original space, they should remain so after projection.

We verify this empirically on the class-imbalanced CIFAR-10 benchmark (see Section \ref{subsection:class-imbalanced-learning}). At each training step, we compute the pairwise cosine similarities between gradient row vectors within each layer, both in the original $n$-dimensional space and in the $d$-dimensional projected space ($d{=}128$). We then measure Pearson correlation between the two sets of cosines and the rate at which projection preserves the sign of each pairwise cosine. To ensure that this property is not an artefact of data abundance, we report results separately for the dominant group (8 classes, full data) and the rare group (2 classes, 10\% data).

\begin{figure}[t]
  \centering
  \includegraphics[width=\linewidth]{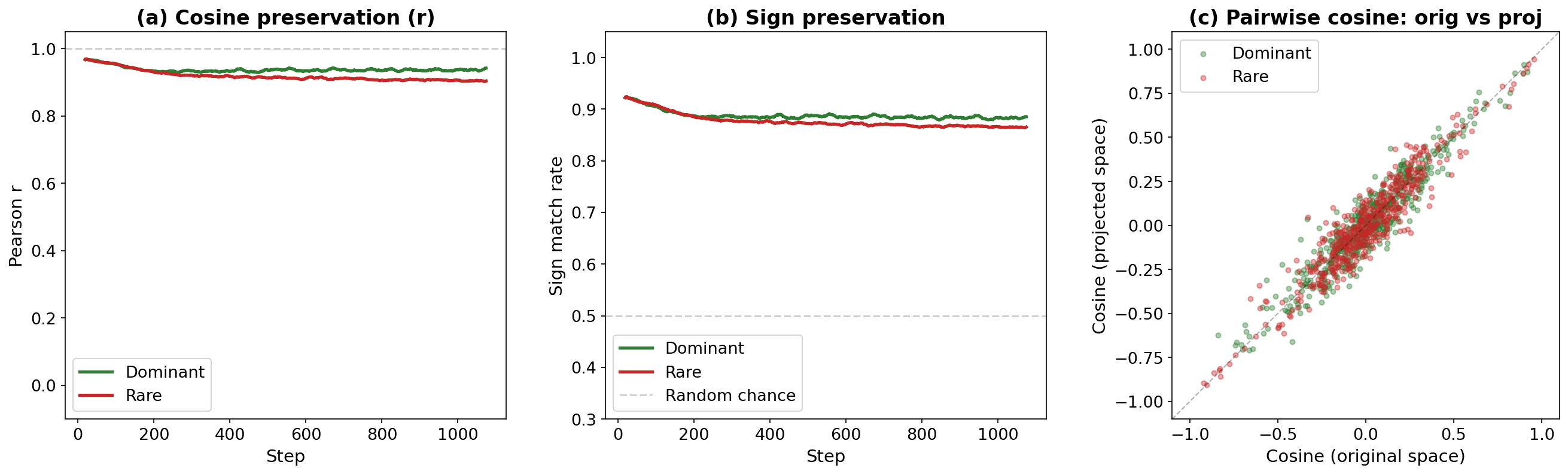}
  \caption{%
    \textbf{Gradient structure preservation under random projection}
    (CIFAR-10, ResNet-18, $d{=}128$).
    \textbf{(a)}~Pearson correlation between pairwise cosine similarities
    in the original and projected spaces remains above 0.9 for both groups.
    \textbf{(b)}~Sign preservation rate stays around 89\%, well above the
    50\% random baseline.
    \textbf{(c)}~Scatter of individual cosine pairs at the final training
    step; both groups cluster tightly around the $y{=}x$ identity line.
    Dominant and rare groups show nearly identical fidelity, confirming that
    projection quality is independent of subgroup sample size.
  }
  \label{fig:projection-fidelity}
\end{figure}

Fig.~\ref{fig:projection-fidelity} summarizes the results. The Rademacher projection preserves pairwise cosine structure with Pearson $r > 0.9$ throughout training for both groups, and sign fidelity remains around 89\%. Crucially, the two groups exhibit nearly identical preservation quality despite a ${\sim}10{\times}$ difference in sample count, confirming that the projection is data-agnostic: it neither favors nor disadvantages any subgroup. The scatter plot further shows that deviations from perfect preservation concentrate near cosine ${\approx}\,0$ (near-orthogonal pairs), where the practical impact on interference detection is minimal. These results justify \method{}'s use of random projection as a faithful dimensionality reduction step that retains the gradient geometry needed for codebook-based conflict management.

\subsection{Short-Term Forgetting in Continual Learning}
\label{subsec:short_term_forgetting_in_cl}

\begin{table}[t]
\centering
\caption{\textbf{Effect of short-term filtering on PACS Domain-IL.}
Results are averaged over three randomly selected domain orders and two seeds.
AP denotes Average Performance, and AF denotes Average Forgetting.}
\label{tab:short_term_filtering_ablation}
\small
\setlength{\tabcolsep}{4.8pt}
\renewcommand{\arraystretch}{1.12}
\resizebox{0.62\linewidth}{!}{%
\begin{tabular}{@{}lccc@{}}
\toprule
\rowcolor{sectionblue}
\textbf{Method}
& \textbf{ST Filtering}
& \textbf{AP (\%)} $\uparrow$
& \textbf{AF (\%)} $\downarrow$ \\
\midrule

\rowcolor{standard}
\method{} w/o ST filtering
& Disabled
& $87.37 \pm 4.23$
& $6.90 \pm 2.48$ \\

\rowcolor{copehl}
\textbf{\textcolor{copegreen}{\method{}}}
& \textcolor{copegreen}{Enabled}
& \textcolor{copegreen}{$\mathbf{87.89 \pm 2.94}$}
& \textcolor{copegreen}{$\mathbf{6.77 \pm 2.31}$} \\

\bottomrule
\end{tabular}%
}
\end{table}

We ablate short-term angular-band filtering while keeping all other components
of \method{} unchanged. As shown in
Table~\ref{tab:short_term_filtering_ablation}, enabling this module improves AP
from \(87.37\%\) to \(87.89\%\) and reduces AF from \(6.90\%\) to \(6.77\%\).
The improvement is modest but consistent, indicating that short-term filtering
helps reduce local interference during training. In PACS Domain-IL, however,
the main retention effect appears to come from long-term centroid protection,
while short-term filtering provides additional stabilization.
%






\subsection{Efficiency Analysis}
\label{subsec:Efficiency_Analysis_appendix}

\subsubsection{Theoretical Algorithm Complexity}

We analyze the per-step cost for one matrix-shaped update block
\(G_t\in\mathbb R^{m\times n}\). Let \(d\) denote the projected memory
dimension, with \(d\ll n\) and typically \(d\ll\min\{m,n\}\). Adam performs
elementwise moment updates and costs \(\mathcal O(mn)\). Muon adds
Newton--Schulz orthogonalization; with \(K_{\mathrm{NS}}\) iterations, its
dominant cost is
\[
\mathcal O\!\left(K_{\mathrm{NS}}mn\min\{m,n\}\right).
\]

\method{} maintains slow and fast orthogonalized streams, giving a
constant-factor overhead over Muon:
\[
\mathcal O\!\left(2K_{\mathrm{NS}}mn\min\{m,n\}\right).
\]
The remaining operations are performed in the projected space
\(\mathbb R^d\). Computing the two projections
\(P_t^{(s)}=O_t^{(s)}\Pi\), \(P_t^{(f)}=O_t^{(f)}\Pi\), and the final lifting
\(\Pi u_{t,i}^{\mathrm{prox}}\) costs \(\mathcal O(3mnd)\) for a dense
projection. The online memory read, including nearest-centroid assignment,
costs \(\mathcal O(mCd)\), where \(C\) is the active codebook size. Short-term
angular-band filtering reuses the codebook similarities and accumulates the
over-aligned and conflicting projected components, costing at most another
\(\mathcal O(mCd)\). Long-term protection compares against
\(K_{\mathrm{cand}}\) sampled frozen centroids and accumulates the protected
component, costing \(\mathcal O(2mK_{\mathrm{cand}}d)\). With
\(M=\Pi^\top\Pi\) precomputed, the proximal solve costs \(\mathcal O(md^2)\).
Thus, the per-step cost is
\[
\mathcal O\!\left(
2K_{\mathrm{NS}}mn\min\{m,n\}
+
3mnd
+
2mCd
+
2mK_{\mathrm{cand}}d
+
md^2
\right).
\]

The projected-memory overhead relative to the dual-stream orthogonalization
term is
\[
\frac{
3mnd+2mCd+2mK_{\mathrm{cand}}d+md^2
}{
2K_{\mathrm{NS}}mn\min\{m,n\}
}
=
\frac{
d(3n+2C+2K_{\mathrm{cand}}+d)
}{
2K_{\mathrm{NS}}n\min\{m,n\}
}.
\]
Hence, when \(d,C,K_{\mathrm{cand}}\ll\min\{m,n\}\), the forgetting-aware memory
correction is lower order than the orthogonalization cost. The dominant overhead
over Muon is therefore the intentional slow--fast dual stream, while the
short-term and long-term memory operations remain comparatively small.

The optimizer-state memory is
\[
\mathcal O(2mn+Cd+K_{\mathrm{lt}}d+d^2),
\]
assuming the random projection is generated implicitly from a seed. The
\(2mn\) term stores the slow and fast momentum states, while \(Cd\),
\(K_{\mathrm{lt}}d\), and \(d^2\) correspond to the active codebook, frozen
centroids, and proximal matrix \(M=\Pi^\top\Pi\), respectively. If \(\Pi\) is
stored explicitly, an additional \(\mathcal O(nd)\) term is required. Thus,
\method{} is heavier than Adam and Muon, but its forgetting-aware memory scales
with the projected dimension \(d\), not the full update dimension.

\subsubsection{Overhead FLOPs}

We next give a concrete FLOP-level estimate to contextualize the overhead. This
estimate counts dominant matrix operations and ignores small elementwise costs,
so it should be interpreted as an order-of-magnitude comparison rather than an
exact wall-clock predictor.

Consider a square block with \(m=n=4096\), projection dimension \(d=512\),
codebook size \(C=384\), sampled frozen centroids \(K_{\mathrm{cand}}=64\), and
\(K_{\mathrm{NS}}=5\). Muon's dominant cost comes from Newton--Schulz
orthogonalization:
\[
K_{\mathrm{NS}}mn\min\{m,n\}
=
5\cdot4096^3
\approx
3.44\times10^{11}.
\]
\method{} applies orthogonalization to both slow and fast streams, giving
\[
2K_{\mathrm{NS}}mn\min\{m,n\}
=
2\cdot5\cdot4096^3
\approx
6.87\times10^{11}.
\]

The projected-space overhead is substantially smaller. Projection and lifting
cost
\[
3mnd
=
3\cdot4096^2\cdot512
\approx
2.58\times10^{10}.
\]
Online codebook lookup costs
\[
mCd
=
4096\cdot384\cdot512
\approx
8.05\times10^8.
\]
Short-term angular-band filtering reuses the codebook similarities and
accumulates the over-aligned and conflicting projected components, costing at
most another
\[
mCd
\approx
8.05\times10^8.
\]
Long-term protection over sampled frozen centroids costs approximately
\[
2mK_{\mathrm{cand}}d
=
2\cdot4096\cdot64\cdot512
\approx
2.68\times10^8,
\]
where the factor \(2\) accounts for alignment computation and protected
component accumulation. Finally, the proximal solve costs
\[
md^2
=
4096\cdot512^2
\approx
1.07\times10^9.
\]

Thus, the additional projected-space operations contribute approximately
\[
2.58\times10^{10}
+
8.05\times10^8
+
8.05\times10^8
+
2.68\times10^8
+
1.07\times10^9
\approx
2.87\times10^{10}.
\]
This is about
\[
\frac{2.87\times10^{10}}{6.87\times10^{11}}
\approx
4.2\%
\]
of the dual-stream orthogonalization cost. Therefore, for \(d=512\), the main
overhead over Muon comes from using two orthogonalized streams, while the
forgetting-aware memory mechanism itself adds only a modest projected-space
cost.

Compared with Adam, \method{} is significantly more expensive because it is not
an elementwise optimizer. Compared with Muon, the overhead is more controlled:
the dominant increase is the slow--fast dual stream, and the short-term and
long-term memory terms scale with \(d\), \(C\), and \(K_{\mathrm{cand}}\), rather
than the full block dimension. This makes the extra cost acceptable when
explicit control of short-term and long-term forgetting is the goal.

\subsubsection{Training Time and Memory Usage}

\begin{table}[t]
\centering
\caption{\textbf{Training time and memory usage across evaluation settings.}
We report wall-clock training time, peak GPU memory allocation, and persistent
method memory. Method memory denotes storage used for replay buffers, stored
models, or centroid memories, separate from transient GPU memory.}
\label{tab:training_time_memory}
\small
\setlength{\tabcolsep}{5.2pt}
\renewcommand{\arraystretch}{1.08}
\resizebox{0.86\linewidth}{!}{%
\begin{tabular}{llccc}
\toprule
\textbf{Benchmark}
& \textbf{Method}
& \textbf{Training Time (mins)}
& \textbf{GPU Memory (GiB)}
& \textbf{Method Memory (MB)} \\
\midrule

\multirow{3}{*}{Class-imbalanced}
& Adam     & 6.51  & 1.85 & -- \\
& Muon      & 7.75  & 1.85 & -- \\
& \method{} & 11.91 & 1.91 & -- \\
\midrule

\multirow{3}{*}{GPT-2 pretraining}
& Adam     & 844$^{*}$ & 79.65 & -- \\
& Muon      & 890$^{*}$ & 79.70 & -- \\
& \method{} & 964$^{*}$ & 79.98 & -- \\
\midrule

\multirow{3}{*}{Class-IL}
& Adam     & 5.51  & 2.10 & 2.35 \\
& Muon      & 6.62  & 2.11 & 2.35 \\
& \method{} & 10.08 & 2.39 & 2.80 \\
\midrule

\multirow{3}{*}{Continual VisualQA}
& AdamW     & 605.0$^{\dagger}$ & 56.08 & $<0.1$ \\
& Muon      & 626.0$^{\dagger}$ & 56.15 & $<0.1$ \\
& \method{} & 857.4$^{\dagger}$ & 59.95 & 36.7 \\
\bottomrule
\end{tabular}
}
\vspace{2pt}

\footnotesize{
$^{*}$Training uses 4 GPUs. 
$^{\dagger}$Training uses 2 GPUs. 
For Continual VisualQA, GPU memory is measured on the Judge task; relative memory differences are similar across tasks.
``--'' indicates no additional persistent method memory, and $<0.1$ MB denotes negligible storage for data-free methods.
}
\end{table}

We evaluate the practical efficiency of \method{} across all experimental
settings. Table~\ref{tab:training_time_memory} reports wall-clock training time,
peak GPU memory allocation, and persistent method memory. The latter isolates
the storage required for forgetting control, such as replay buffers, stored
models, or centroid memories, from transient GPU memory used during training.

Overall, \method{} introduces only a modest increase in peak GPU memory across
settings, while requiring a small amount of persistent memory for its centroid
state. The overhead is more visible in smaller vision experiments, where the
baseline runtime is short, but remains moderate relative to the full training
cost in large-scale GPT-2 pretraining and Continual VisualQA. Notably, the
persistent memory of \method{} remains compact: it avoids storing replay data or
previous model snapshots, and instead maintains only lightweight directional
centroid memories for forgetting control.

\subsection{Expert Representation Geometry in Continual VisualQA}
\label{sec:appendix_cl_implications}

\begin{figure}[htbp]
    \centering
    \begin{subfigure}[b]{0.48\textwidth}
        \centering
        \includegraphics[width=\textwidth]{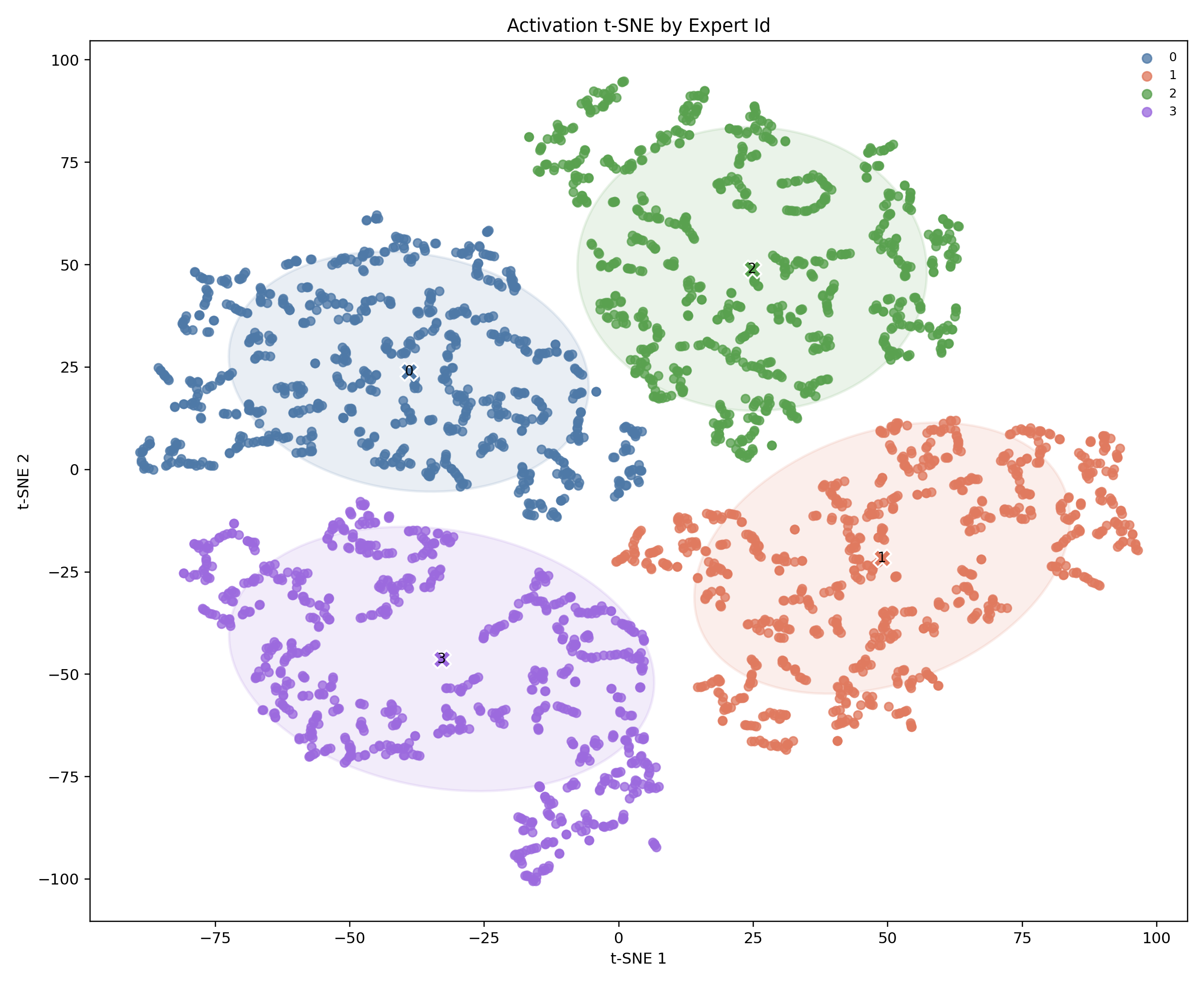}
        \caption{\method{}: balanced expert variance}
        \label{fig:tsne_balanced}
    \end{subfigure}
    \hfill
    \begin{subfigure}[b]{0.48\textwidth}
        \centering
        \includegraphics[width=\textwidth]{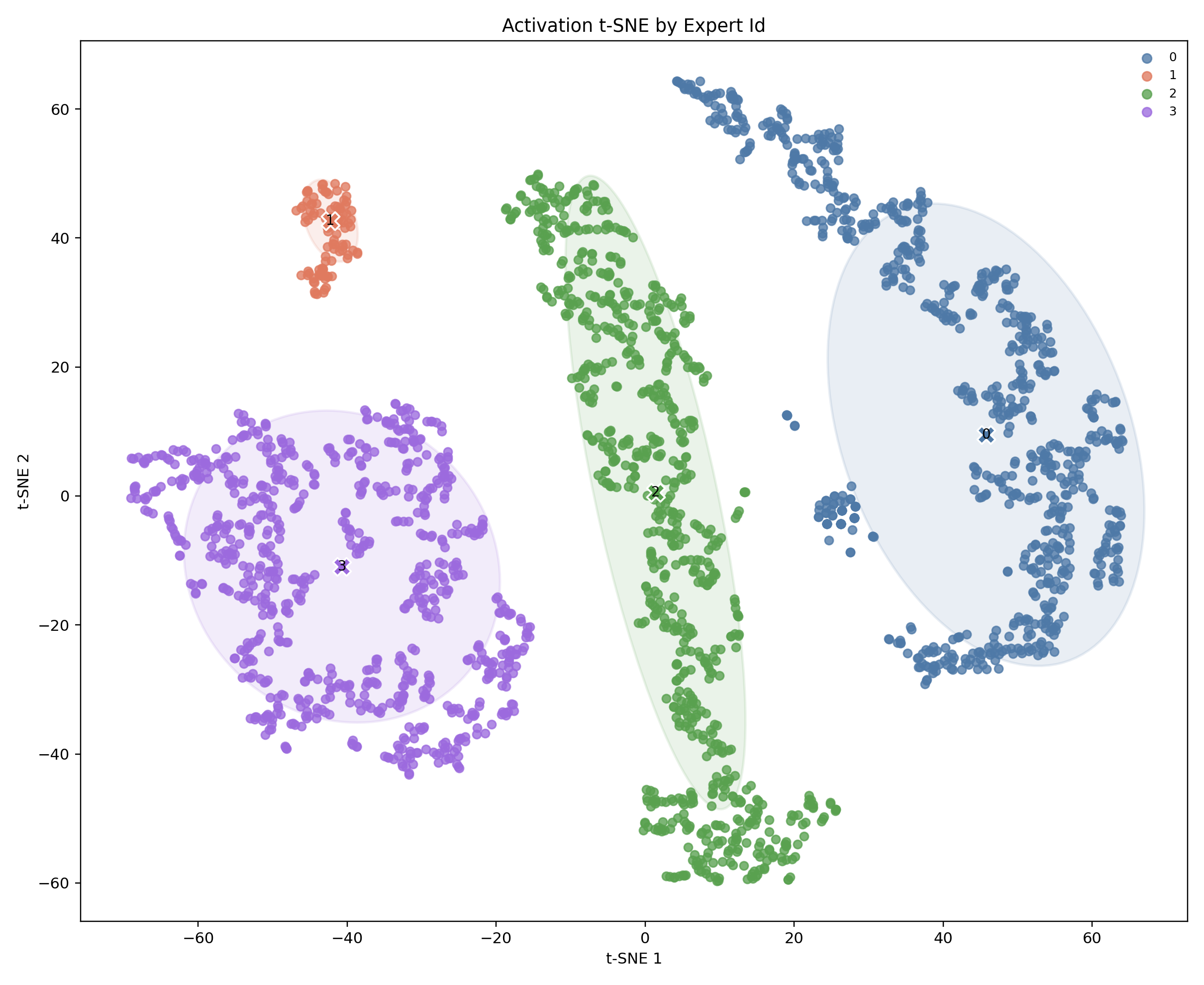}
        \caption{AdamW: variance imbalance and collapse}
        \label{fig:tsne_imbalance}
    \end{subfigure}
    
    \caption{\textbf{t-SNE visualization of expert activations in continual VisualQA.}
    Both plots use the same number of sampled points per expert, isolating representation geometry from routing-load differences. \textbf{(a)} \method{} maintains balanced within-expert variance and clear expert separation, suggesting a healthier distribution of task-specific representations. \textbf{(b)} AdamW leads to representation imbalance, where one expert collapses into a narrow region while others cover scattered high-variance features, increasing susceptibility to forgetting under sequential task learning.}
    \label{fig:tsne_expert_activations}
\end{figure}

We further analyze why \method{} outperforms other optimizers in continual VisualQA by visualizing expert activations after training on a sequence of tasks. Figure~\ref{fig:tsne_expert_activations} compares the representation geometry induced by \method{} and AdamW. To make the comparison independent of routing frequency, both plots use the same number of sampled points per expert.

AdamW produces a highly imbalanced representation space. As shown in Figure~\ref{fig:tsne_imbalance}(b), one expert collapses into a narrow region, while the remaining experts cover broader and more scattered feature regions. This suggests uneven expert utilization at the representation level: collapsed experts become overly specialized and lose plasticity, whereas high-variance experts must absorb heterogeneous task patterns. In a continual learning setting, such imbalance is harmful because new tasks are likely to overwrite the broadly activated experts, while collapsed experts contribute little to future adaptation.

In contrast, \method{} yields a more balanced expert geometry, as shown in Figure~\ref{fig:tsne_balanced}(a). The expert clusters remain well separated while preserving comparable within-expert diversity. This indicates that \method{} distributes task-related variation more evenly across experts, avoiding both feature collapse and excessive dispersion. Such balanced variance provides a more favorable stability--plasticity trade-off: each expert retains enough capacity to adapt to new VisualQA tasks without concentrating the burden of continual updates on only a small subset of experts.

\section{Case Study and Qualitative Analysis}

In this section, we present a qualitative comparison between the baseline LoRA-MoE trained continually with AdamW and our proposed \method{} model across various VisualQA scenarios. We selected four representative cases to demonstrate \method{}'s superiority in fine-grained recognition, scene understanding, counting, and reasoning.

\begin{figure}[htbp]
    \centering
    \begin{subfigure}[b]{0.48\textwidth}
        \centering
        \includegraphics[page=2, width=\textwidth]{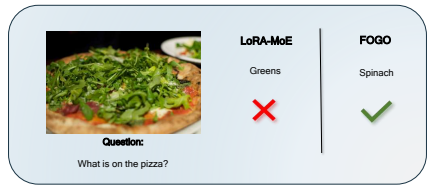}
        \caption{Fine-grained Recognition}
        \label{fig:case_pizza}
    \end{subfigure}
    \hfill
    \begin{subfigure}[b]{0.48\textwidth}
        \centering
        \includegraphics[page=3, width=\textwidth]{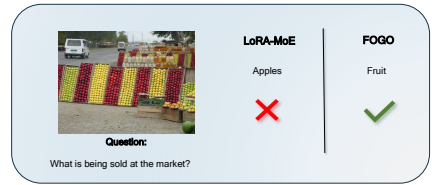}
        \caption{Scene Context Understanding}
        \label{fig:case_market}
    \end{subfigure}

    \vspace{1em}

    \begin{subfigure}[b]{0.48\textwidth}
        \centering
        \includegraphics[page=4, width=\textwidth]{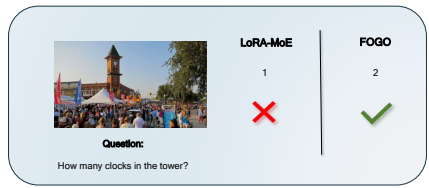}
        \caption{Visual Counting}
        \label{fig:case_clocks}
    \end{subfigure}
    \hfill
    \begin{subfigure}[b]{0.48\textwidth}
        \centering
        \includegraphics[page=5, width=\textwidth]{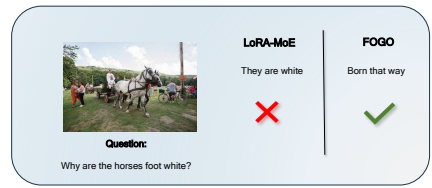}
        \caption{Commonsense Reasoning}
        \label{fig:case_horses}
    \end{subfigure}

    \caption{Qualitative comparison between LoRA-MoE and FOGO. FOGO consistently provides more precise, context-aware, and logically sound answers.}
    \label{fig:qualitative_comparison}
\end{figure}

As illustrated in Figure \ref{fig:qualitative_comparison}, FOGO outperforms the baseline in several key dimensions:
\begin{itemize}
    \item \textbf{Specificity in Recognition:} In the pizza example (Fig. \ref{fig:case_pizza}), while LoRA-MoE provides a generic category (``Greens''), FOGO correctly identifies the specific ingredient (``Spinach''), showcasing superior fine-grained visual capabilities.
    \item \textbf{Holistic Scene Interpretation:} For the market scene (Fig. \ref{fig:case_market}), FOGO delivers a comprehensive summary (``Fruit'') rather than focusing on a single item (``Apples''), reflecting a better grasp of global context.
    \item \textbf{Numerical Precision:} FOGO accurately counts the objects in the tower (Fig. \ref{fig:case_clocks}), whereas LoRA-MoE fails to detect all relevant targets in the counting task.
    \item \textbf{Logical Reasoning:} In response to why the horses' feet are white (Fig. \ref{fig:case_horses}), FOGO moves beyond tautological descriptions (``They are white'') to provide a commonsense-based causal explanation (``Born that way'').
\end{itemize}

\section{Limitations and Future Work}
\label{sec:limitation_appendix}

\method{} requires {choosing} a codebook size and {a} projection dimension; we {set the projection dimension from the Johnson--Lindenstrauss bound} and find {these defaults} stable across all our settings, though they may benefit from adaptive tuning for very long task sequences. The codebook capacity is currently fixed and cannot yet grow or compress in response to changing task complexity. Additionally, spectral orthogonalization and codebook maintenance introduce modest computational overhead compared to Adam {(Appendix~\ref{subsec:Efficiency_Analysis_appendix})}; we find this acceptable given the consistent gains, but further engineering{, such as reduced-frequency codebook updates,} could narrow the gap. More broadly, we believe a complete solution to forgetting requires joint effort across optimization, architecture, and data strategy; \method{} addresses the first axis and is designed to complement advances in the other two. Promising future directions include {adaptive codebook sizing and projection dimensions,} and extension to fully boundary-free continual learning.

\newpage

\providecommand{\rev}[1]{\textcolor{red}{#1}}%
\section*{NeurIPS Paper Checklist}
\begin{enumerate}

\item {\bf Claims}
    \item[] Question: Do the main claims made in the abstract and introduction accurately reflect the paper's contributions and scope?
    \item[] Answer: \answerYes{} 
    \item[] Justification: The paper's contributions are claimed in the abstract and section~\ref{sec:Intro}.

\item {\bf Limitations}
    \item[] Question: Does the paper discuss the limitations of the work performed by the authors?
    \item[] Answer: \answerYes{} 
    \item[] Justification: The limitations are discussed in Appendix~\ref{sec:limitation_appendix}

\item {\bf Theory assumptions and proofs}
    \item[] Question: For each theoretical result, does the paper provide the full set of assumptions and a complete (and correct) proof?
    \item[] Answer:  \answerYes{} 
    \item[] Justification: Complete assumptions and proofs for all theoretical results are provided in Appendix~\ref{sect:theoretical_results_appendix}.

    \item {\bf Experimental result reproducibility}
    \item[] Question: Does the paper fully disclose all the information needed to reproduce the main experimental results of the paper to the extent that it affects the main claims and/or conclusions of the paper (regardless of whether the code and data are provided or not)?
    \item[] Answer: \answerYes{} 
    \item[] Justification: Full experimental settings are described in Section~\ref{sec:Experiments}, with additional implementation details in Appendix~\ref{sec:more_exps_settings_appendix}. We also {release our code}.

\item {\bf Open access to data and code}
    \item[] Question: Does the paper provide open access to the data and code, with sufficient instructions to faithfully reproduce the main experimental results, as described in supplemental material?
    \item[] Answer: \answerYes{} 
    \item[] Justification: Our code is {publicly} available at the URL provided in the abstract.

\item {\bf Experimental setting/details}
    \item[] Question: Does the paper specify all the training and test details (e.g., data splits, hyperparameters, how they were chosen, type of optimizer) necessary to understand the results?
    \item[] Answer: \answerYes{} 
    \item[] Justification: Please refer to Appendix~\ref{sec:more_exps_settings_appendix}.

\item {\bf Experiment statistical significance}
    \item[] Question: Does the paper report error bars suitably and correctly defined or other appropriate information about the statistical significance of the experiments?
    \item[] Answer: \answerYes{} 
    \item[] Justification:  All results are reported with mean and standard deviation across multiple seeds.

\item {\bf Experiments compute resources}
    \item[] Question: For each experiment, does the paper provide sufficient information on the computer resources (type of compute workers, memory, time of execution) needed to reproduce the experiments?
    \item[] Answer: \answerYes{} 
    \item[] Justification: Please refer to Appendix~\ref{sec:more_exps_settings_appendix} and Appendix~\ref{subsec:Efficiency_Analysis_appendix}.
    
\item {\bf Code of ethics}
    \item[] Question: Does the research conducted in the paper conform, in every respect, with the NeurIPS Code of Ethics \url{https://neurips.cc/public/EthicsGuidelines}?
    \item[] Answer: \answerYes{} 
    \item[] Justification: {The research conforms, in every respect, with the NeurIPS Code of Ethics.}

\item {\bf Broader impacts}
    \item[] Question: Does the paper discuss both potential positive societal impacts and negative societal impacts of the work performed?
    \item[] Answer: \answerYes{} 
    \item[] Justification: \method{} advances continual learning by improving model retention and adaptation under non-stationary data streams. This may support more reliable and resource-efficient machine learning systems. Since the work is methodological and evaluated on standard benchmarks, we do not anticipate direct negative societal impacts beyond general risks associated with deploying machine learning models.
    
\item {\bf Safeguards}
    \item[] Question: Does the paper describe safeguards that have been put in place for responsible release of data or models that have a high risk for misuse (e.g., pre-trained language models, image generators, or scraped datasets)?
    \item[] Answer: \answerNA{} 
    \item[] Justification: the paper poses no such risks.

\item {\bf Licenses for existing assets}
    \item[] Question: Are the creators or original owners of assets (e.g., code, data, models), used in the paper, properly credited and are the license and terms of use explicitly mentioned and properly respected?
    \item[] Answer: \answerYes{} 
    \item[] Justification: {All datasets, pretrained models, and code used in this work are credited by citation and used under their original licenses and terms of use.}

\item {\bf New assets}
    \item[] Question: Are new assets introduced in the paper well documented and is the documentation provided alongside the assets?
    \item[] Answer: \answerYes{} 
    \item[] Justification: {We release our code, with usage instructions, at the URL provided in the abstract.}

\item {\bf Crowdsourcing and research with human subjects}
    \item[] Question: For crowdsourcing experiments and research with human subjects, does the paper include the full text of instructions given to participants and screenshots, if applicable, as well as details about compensation (if any)? 
    \item[] Answer: \answerNA{} 
    \item[] Justification: The paper does not involve crowdsourcing nor research with human subjects.

\item {\bf Institutional review board (IRB) approvals or equivalent for research with human subjects}
    \item[] Question: Does the paper describe potential risks incurred by study participants, whether such risks were disclosed to the subjects, and whether Institutional Review Board (IRB) approvals (or an equivalent approval/review based on the requirements of your country or institution) were obtained?
    \item[] Answer: \answerNA{} 
    \item[] Justification: The paper does not involve crowdsourcing nor research with human subjects

\item {\bf Declaration of LLM usage}
    \item[] Question: Does the paper describe the usage of LLMs if it is an important, original, or non-standard component of the core methods in this research? Note that if the LLM is used only for writing, editing, or formatting purposes and does \emph{not} impact the core methodology, scientific rigor, or originality of the research, declaration is not required.
    \item[] Answer: \answerNA{} 
    \item[] Justification: The core method development in this research does not involve LLMs as any
important, original, or non-standard components. We use LLMs only for routine text editing and polishing.

\end{enumerate}

\end{document}